\documentclass[10pt,doublespacing]{elsart}
\usepackage{latexsym,amssymb,amsmath,graphicx,epsfig,cite,bbm,float,epsf,graphics}
\usepackage{algorithm}
\usepackage{algpseudocode}
\usepackage{tabularx}
\usepackage{url}
\usepackage[titletoc,title]{appendix}
\usepackage{multicol}

\journal{Signal Processing}

\makeatletter
\renewcommand{\ALG@beginalgorithmic}{\scriptsize}
\makeatother

\newcolumntype{C}{>{\centering\arraybackslash}X}

\DeclareMathOperator*{\argmin}{arg\,min}

\renewcommand{\vec}[1]{\mbox{\boldmath${#1}$}}

\newcommand{\ei}{\end{itemize}}
\newcommand{\bi}{\begin{itemize}}

\newcommand{\MB}{\left[\begin{array}}
\newcommand{\ME}{\end{array}\right]}

\makeatletter

\newcommand{\Rmnum}[1]{\expandafter\@slowromancap\romannumeral #1@}
\makeatother

\def\ninept{\def\baselinestretch{1.5}}
\ninept

\begin{document}
\begin{frontmatter}

\title{Highly Efficient Hierarchical Online Nonlinear Regression Using Second Order Methods}
\vspace{-0.5in}
\author[1]{Burak C. Civek\corauthref{cor}},
\ead{civek@ee.bilkent.edu.tr}
\author[2]{Ibrahim Delibalta},
\ead{ibrahim.delibalta@turktelekom.com.tr}
\author[1]{Suleyman S. Kozat}
\corauth[cor]{Corresponding author.}
\ead{kozat@ee.bilkent.edu.tr}
\address[1]{Department of Electrical and Electronics Engineering, Bilkent University, Ankara, Turkey.}
\address[2]{Turk Telekom Communications Services Inc., Istanbul, Turkey.}
\vspace{-0.3in}
\begin{abstract}
We introduce highly efficient online nonlinear regression algorithms that are suitable for real life applications. We process the data in a truly online manner such that no storage is needed, i.e., the data is discarded after being used. For nonlinear modeling we use a hierarchical piecewise linear approach based on the notion of decision trees where the space of the regressor vectors is adaptively partitioned based on the performance. As the first time in the literature, we learn both the piecewise linear partitioning of the regressor space as well as the linear models in each region using highly effective second order methods, i.e., Newton-Raphson Methods. Hence, we avoid the well known over fitting issues by using piecewise linear models, however, since both the region boundaries as well as the linear models in each region are trained using the second order methods, we achieve substantial performance compared to the state of the art. We demonstrate our gains over the well known benchmark data sets and provide performance results in an individual sequence manner guaranteed to hold without any statistical assumptions. Hence, the introduced algorithms address computational complexity issues widely encountered in real life applications while providing superior guaranteed performance in a strong deterministic sense.
\end{abstract}
\end{frontmatter} 
\section{Introduction} 
Recent developments in information technologies, intelligent use of mobile devices and Internet have bolstered the capacity and capabilities of data acquisition systems beyond expectation \cite{SP1, SP2, Kozattree, sensor2, KDE3, bigdata1, bigdata2, bigdata3}. Today, many sources of information from shares on social networks to blogs, from intelligent device activities to security camera recordings are easily accessible. Efficient and effective processing of this data can significantly improve the performance of many signal processing and machine learning algorithms \cite{sayedbook, bigdata7, Moon1}. In this paper, we investigate the nonlinear regression problem that is one of the most important topics in the machine learning and signal processing literatures. This problem arises in several different applications such as signal modeling \cite{Singer3, Hero}, financial market \cite{financial} and trend analyses \cite{trend}, intrusion detection \cite{intrusion} and recommendation \cite{recommendation}. However, traditional regression techniques show less than adequate performance in real-life applications having big data since (1) data acquired from diverse sources are too large in size to be efficiently processed or stored by conventional signal processing and machine learning methods \cite{bigdata2, bigdata3, cesabook}; (2) the performance of the conventional methods is further impaired by the highly variable properties, structure and quality of  data acquired at high speeds \cite{bigdata2, bigdata3}.\par
In this context, to accommodate these problems, we introduce online regression algorithms that process the data in an online manner, i.e., instantly, without any storage, and then discard the data after using and learning \cite{Singer2, cesabook}. Hence our methods can constantly adapt to the changing statistics or quality of the data so that they can be robust and prone to variations and uncertainties \cite{kozat, Singer2, yilmaz}. From a unified point of view, in such problems, we sequentially observe a real valued sequence vector sequence $\vec{x}_1,\vec{x}_2, \ldots$ and produce a decision (or an action) $d_t$ at each time $t$ based on the past $\vec{x}_1, \vec{x}_2, \ldots, \vec{x}_t$.  After  the desired output $d_t$ is revealed, we suffer a loss and our goal is to minimize the accumulated (and possibly weighted) loss as much as possible while using a limited amount of information from the past. \par
To this end, for nonlinear regression, we use a hierarchical piecewise linear model based on the notion of decision trees, where the space of the regressor vectors, $\vec{x}_1,\vec{x}_2, \ldots$, is adaptively partitioned and continuously optimized in order to enhance the performance \cite{RPTrees, Helmbold, Hero}. We note that the piecewise linear models are extensively used in the signal processing literature to mitigate the overtraining issues that arise because of using nonlinear models \cite{Hero}. However their performance in real life applications are less than adequate since their successful application highly depends on the accurate selection of the piecewise regions that correctly model the underlying data \cite{CTW}. Clearly, such a goal is impossible in an online setting since either the best partition is not known, i.e., the data arrives sequentially, or in real life applications the statistics of the data and the best selection of the regions change in time. To this end, as the first time in the literature, we learn both the piecewise linear partitioning of the regressor space as well as the linear models in each region using highly effective second order methods, i.e., Newton-Raphson Methods \cite{optimization}. Hence, we avoid the well known over fitting issues by using piecewise linear models, moreover, since both the region boundaries as well as the linear models in each region are trained using the second order methods we achieve substantial performance compared to the state of the art \cite{optimization}. We demonstrate our gains over the well known benchmark data sets extensively used in the machine learning literature. We also provide theoretical performance results in an individual sequence manner that are guaranteed to hold without any statistical assumptions \cite{cesabook}. In this sense, the introduced algorithms address computational complexity issues widely encountered in real life applications while providing superior guaranteed performance in a strong deterministic sense.\par
In adaptive signal processing literature, there exist methods which develop an approach based on weighted averaging of all possible models of a tree based partitioning instead of solely relying on a particular piecewise linear model \cite{Helmbold, CTW}. These methods use the entire partitions of the regressor space and implement a full binary tree to form an online piecewise linear regressor. Such approaches are confirmed to lessen the bias variance trade off in a deterministic framework \cite{Helmbold,CTW}. However, these methods do not update the corresponding partitioning of the regressor space based on the upcoming data. One such example is that the recursive dyadic partitioning, which partitions the regressor space using separation functions that are required to be parallel to the axes \cite{Kolaczyk}. Moreover, these methods usually do not provide a theoretical justification for the weighting of the models, even if there exist inspirations from information theoretic deliberations \cite{Willems}. For instance, there is an algorithmic concern on the definitions of both the exponentially weighted performance measure and the "universal weighting" coefficients \cite{CTW,Singer1,Singer2,linder2} instead of a complete theoretical justifications (except the universal bounds). Specifically, these methods are constructed in such a way that there is a significant correlation between the weighting coefficients, algorithmic parameters and their performance, i.e., one should adjust these parameters to the specific application for successful process \cite{CTW}. Besides these approaches, there exists an algorithm providing adaptive tree structure for the partitions, e.g., the Decision Adaptive Tree (DAT) \cite{DAT}. The DAT produces the final estimate using the weighted average of the outcomes of all possible subtrees, which results in a computational complexity of $O(m4^d)$, where $m$ is the data dimension and $d$ represents the depth. However, this would affect the computational efficiency adversely for the cases involving highly nonlinear structures. In this work, we propose a different approach that avoids combining the prediction of each subtrees and offers a computational complexity of $O(m^22^d)$. Hence, we achieve an algorithm that is more efficient and effective for the cases involving higher nonlinearities, whereas the DAT is more feasible when the data dimension is quite high. Moreover, we illustrate in our experiments that our algorithm requires less number of data samples to capture the underlying data structure. Overall, the proposed methods are completely generic such that they are capable of incorporating all Recursive Dyadic, Random Projection (RP) and $k$-d trees in their framework, e.g., we initialize the partitioning process by using the RP trees and adaptively learn the complete structure of the tree based on the data progress to minimize the final error. \par
In Section \ref{sec:prob}, we first present the main framework for nonlinear regression and piecewise linear modeling. In Section \ref{sec:solution}, we propose three algorithms with regressor space partitioning and present guaranteed upper bounds on the performances. These algorithms adaptively learn the partitioning structure, region boundaries and region regressors to minimize the final regression error.  We then demonstrate
the performance of our algorithms through widely used benchmark data sets in Section
\ref{sec:simulations}. We then finalize our paper with concluding
remarks.

\section{Problem Description}\label{sec:prob}
In this paper, all vectors are column vectors and represented by lower case boldface letters. For matrices, we use upper case boldface letters. The $ \ell^{2} $-norm of a vector $\vec{x}$ is given by $\parallel$$\vec{x}$$\parallel = \sqrt{\vec{x}^{T}\vec{x}}$ where $\vec{x}^{T}$ denotes the ordinary transpose. The identity matrix with $n \times n$ dimension is represented by $\vec{I}_{n} $.\par 
We work in an online setting, where we estimate a data sequence $y_{t} \in \mathbbm{R}$ at time $t\geq1$ using the corresponding observed feature vector $\vec{x}_{t} \in \mathbbm{R}^{m}$ and then discard $\vec{x}_{t}$ without any storage. Our goal is to sequentially estimate $y_{t}$ using $\vec{x}_{t}$ as
\[ \hat{y}_{t} = f_{t}(\vec{x}_{t}) \]
where $f_{t}(\cdot)$ is a function of past observations. In this work, we use nonlinear functions to model $y_{t}$, since in most real life applications, linear regressors are inadequate to successively model the intrinsic relation between the feature vector $\vec{x}_t$ and the desired data $y_t$ \cite{linearmodeling}. Different from linear regressors, nonlinear functions are quite powerful and usually overfit in most real life cases \cite{machinelearning}. To this end, we choose piecewise linear functions due to their capability of approximating most nonlinear models \cite{PL}. In order to construct a piecewise linear model, we partition the space of regressor vectors into $K$ distinct $m$-dimensional regions $S_{k}^{m}$, where $\bigcup_{k = 1}^{K}S_{k}^{m} = \mathbbm{R}^{m}$ and $S_{i}^{m} \cap S_{j}^{m} = \emptyset$ when $i \neq j$. In each region, we use a linear regressor, i.e., $\hat{y}_{t,i} = \vec{w}_{t,i}^{T}\vec{x}_{t} + c_{t,i}$, where $\vec{w}_{t,i}$ is the linear regression vector, $c_{t,i}$ is the offset and $\hat{y}_{t,i}$ is the estimate corresponding to the $i^{th}$ region. We represent $\hat{y}_{t,i}$ in a more compact form as $\hat{y}_{t,i} = \vec{w}_{t,i}^{T}\vec{x}_{t}$, by including a bias term into each weight vector $\vec{w}_{t,i}$ and increasing the dimension of the space by 1, where the last entry of $\vec{x}_t$ is always set to 1. \par
\begin{figure}[t]
	\centering
	\includegraphics[scale=0.35]{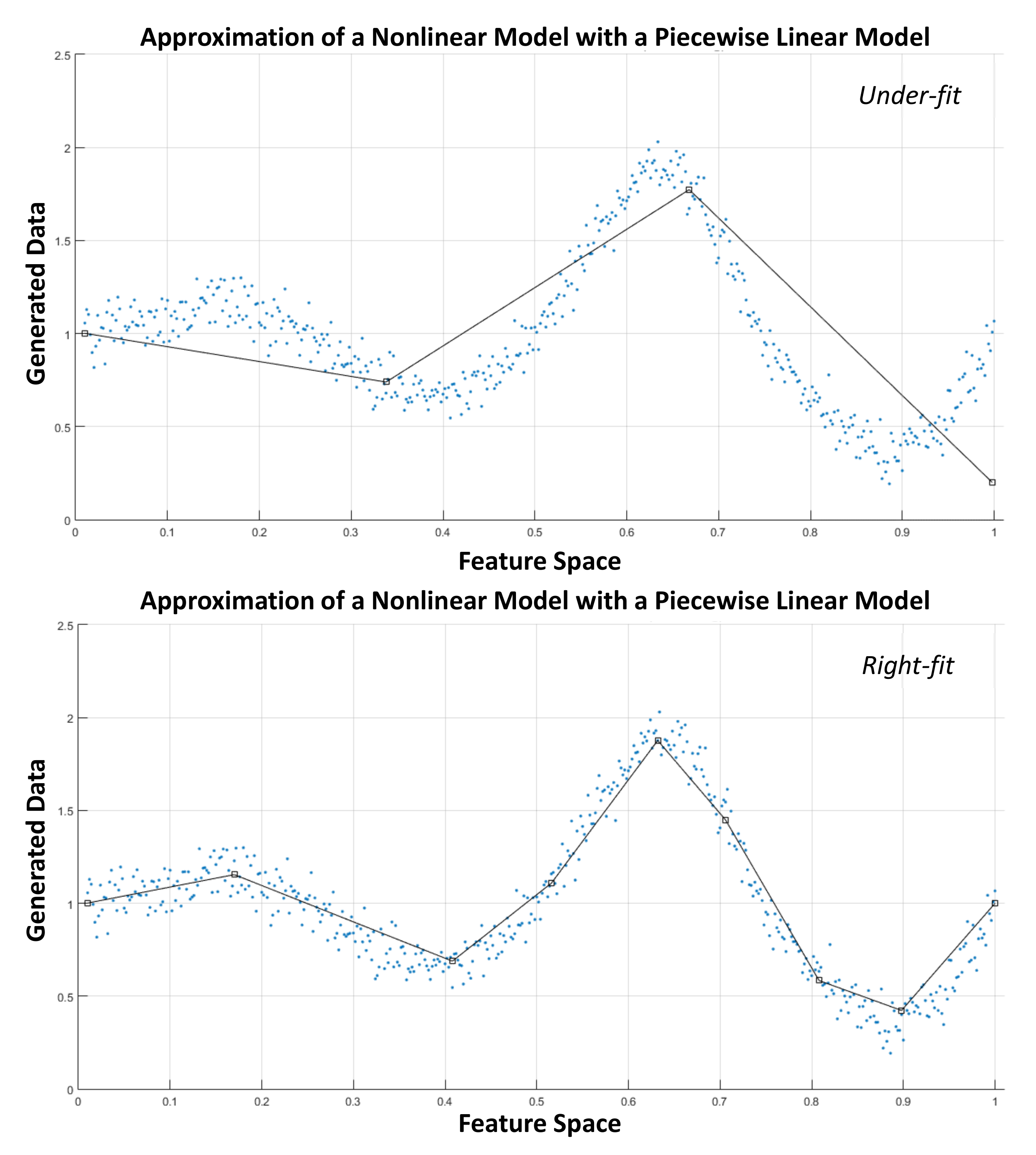} 
	\caption{In the upper plot, we represent an inadequate approximation of a piecewise linear model. In the lower plot, we represent a successive modeling with sufficiently partitioned regression space. \label{fig:PLRegressor}}
\end{figure}
To clarify the framework, in Fig. \ref{fig:PLRegressor}, we present a one dimensional regression problem, where we generate the data sequence using the nonlinear model \[ y_{t} = \exp\bigg(x_{t} \sin(4\pi x_{t})\bigg) + \nu_{t}, \] where $x_{t}$ is a sample function from an i.i.d. standard uniform random process and $\nu_{t}$ has normal distribution with zero mean and 0.1 variance. Here, we demonstrate two different cases to emphasize the difficulties in piecewise linear modeling. For the case given in the upper plot, we partition the regression space into three regions and fit linear regressors to each partition. However, this construction does not approximate the given nonlinear model well enough since the underlying partition does not match exactly to the data. In order to better model the generated data, we use the second model as shown in the lower plot, where we have eight regions particularly selected according to the distribution of the data points. As the two cases signified in Fig. \ref{fig:PLRegressor} imply, there are two major problems when using piecewise linear models. The first one is to determine the piecewise regions properly. Randomly selecting the partitions causes inadequately approximating models as indicated in the underfitting case on the top of Fig. \ref{fig:PLRegressor} \cite{RPTrees}. The second problem is to find out the linear model that best fits the data in each distinct region in a sequential manner \cite{CTW}. In this paper, we solve both of these problems using highly effective and completely adaptive second order piecewise linear regressors.\par
In order to have a measure on how well the determined piecewise linear model fits the data, we use instantaneous squared loss, i.e.,
$ e_{t}^{2} = (y_{t} - \hat{y}_{t})^{2} $ as our cost function. Our goal is to specify the partitions and the corresponding linear regressors at each iteration such that the total regression error is minimized. Suppose $\vec{w}_{n}^{*}$ represents the optimal fixed weight for a particular region after $n$ iteration, i.e.,
\[ \vec{w}_{n}^{*} =  \argmin_{\vec{w}}\sum\limits_{t = 1}^{n}e_{t}^{2}(\vec{w}).\] 
Hence, we would achieve the minimum possible regression error, if we have been considering $\vec{w}_{n}^{*}$ as the fixed linear regressor weight up to the current iteration, $n$. However, we do not process batch data sets, since the framework is online, and thus, cannot know the optimal weight beforehand \cite{cesabook}. This lack of information motivates us to implement an algorithm such that we achieve an error rate as close as the possible minimum after $n$ iteration. At this point, we define the regret of an algorithm to measure how much the total error diverges from the possible minimum achieved by $\vec{w}_{n}^{*}$, i.e.,
\[ \mathrm{Regret}(A) = \sum\limits_{t=1}^{n}e_{t}^{2}(\vec{w}_{t}) - \sum\limits_{t=1}^{n}e_{t}^{2}(\vec{w}_{n}^{*}), \]
where $A$ denotes the algorithm to adjust $\vec{w}_{t}$ at each iteration. Eventually, we consider the regret criterion to measure the modeling performance of the designated piecewise linear model and aim to attain a low regret \cite{cesabook}. \par
In the following section, we propose three different algorithms to sufficiently model the intrinsic relation between the data sequence $y_{t}$ and the linear regressor vectors. In each algorithm, we use piecewise linear models, where we partition the space of regressor vectors by using linear separation functions and assign a linear regressor to each partition. At this point, we also need to emphasize that we propose generic algorithms for nonlinear modeling. Even though we employ linear models in each partition, it is also possible to use, for example, spline modeling within the presented settings. This selection would cause additional update operations with minor changes for the higher order terms. Therefore, the proposed approaches can be implemented by using any other function that is differentiable without a significant difference in the algorithm, hence, they are universal in terms of the possible selection of functions. Overall, the presented algorithms ensure highly efficient and effective learning performance, since we perform second order update methods, e.g. Online Newton Step \cite{Hazan}, for training of the region boundaries and the linear models.

\section{Highly Efficient Tree Based Sequential Piecewise Linear Predictors}\label{sec:solution}

In this section, we introduce three highly effective algorithms constructed by piecewise linear models. The presented algorithms provide efficient learning even for highly nonlinear data models. Moreover, continuous updating based on the upcoming data ensures our algorithms to achieve outstanding performance for online frameworks. Furthermore, we also provide a regret analysis for the introduced algorithms demonstrating strong guaranteed performance.\par
\begin{figure}[t]
	\centering
	\includegraphics[scale=0.45]{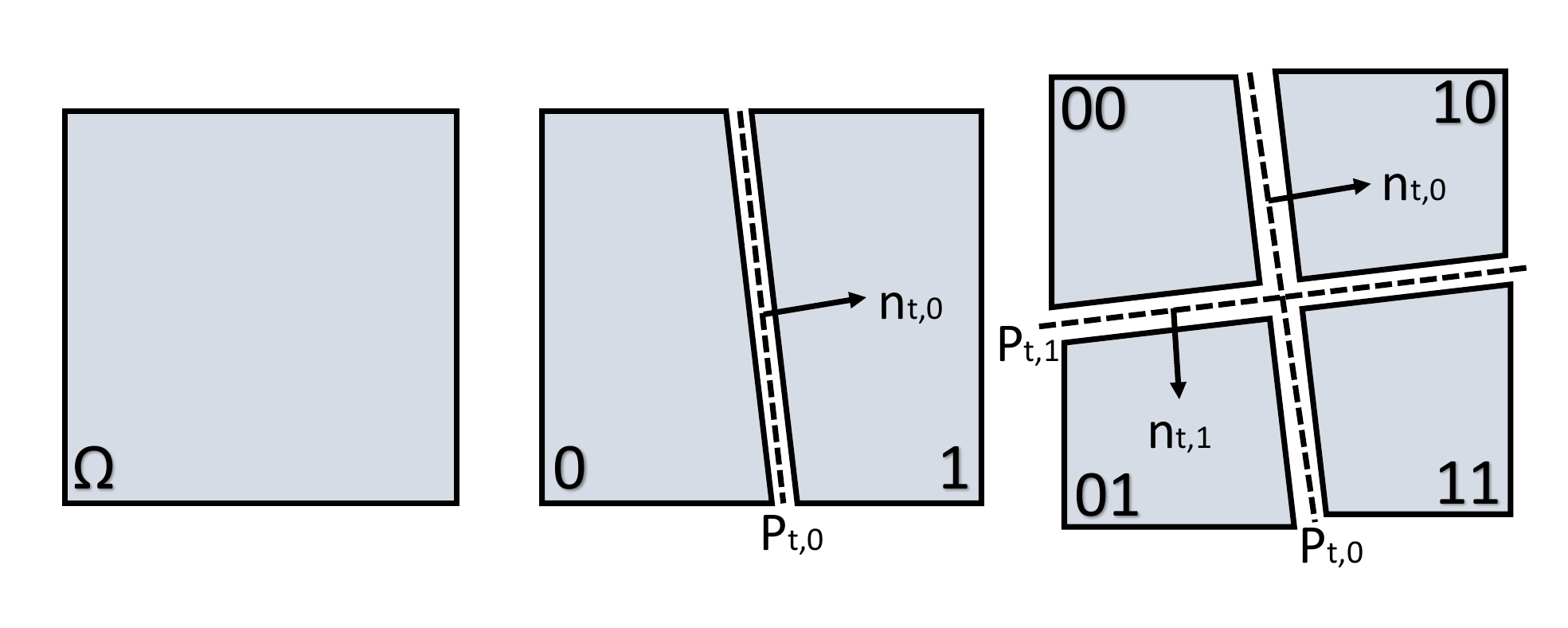}
	\caption{Straight Partitioning of The Regression Space \label{fig:straightpartition}}
\end{figure}
\begin{figure}[t]
	\centering
	\includegraphics[scale=0.3]{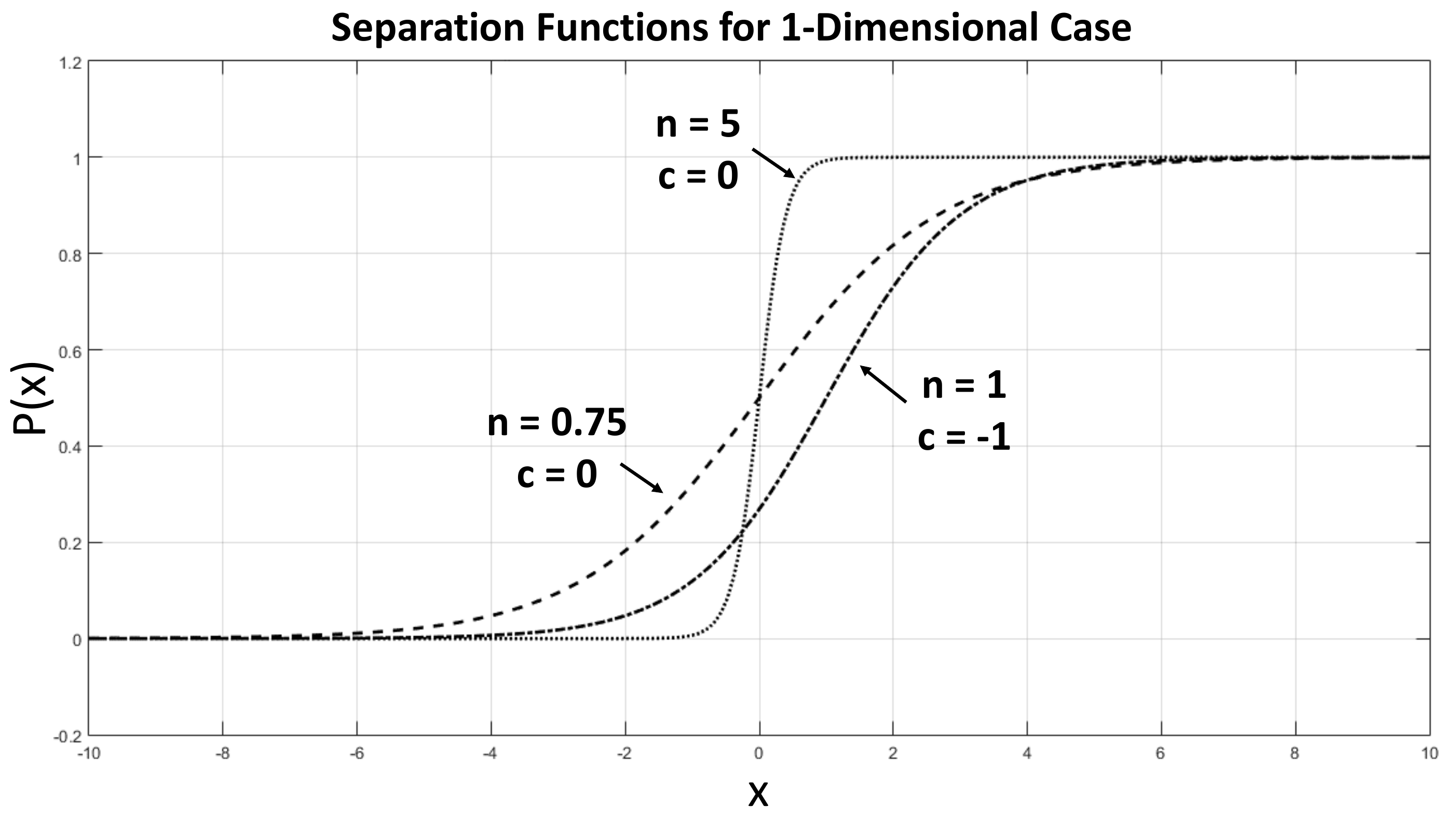} 
	\caption{Separation Functions for 1-Dimensional Case where $\{n = 5,c = 0\}$, $\{n = 0.75,c = 0\}$ and $\{n = 1,c = -1\}$. Parameter $n$ specifies the sharpness, as $c$ determines the position or the offset on the $x$-axis. \label{fig:sepfunction}}
\end{figure}

There exist two essential problems of piecewise linear modeling. The first significant issue is to determine how to partition the regressor space. We carry out the partitioning process using linear separation functions. We specify the separation functions as hyperplanes, which are $(m-1)$-dimensional subspaces of $m$-dimensional regression space and identified by their normal vectors as shown in Fig. \ref{fig:straightpartition}. To get a highly versatile and data adaptive partitioning, we also train the region boundaries by updating corresponding normal vectors. We denote the separation functions as $p_{t,k}$ and the normal vectors as $\vec{n}_{t,k}$ where $k$ is the region label as we demonstrate in Fig. \ref{fig:straightpartition}. In order to adaptively train the region boundaries, we use differentiable functions as the separation functions instead of hard separation boundaries as seen in Fig. \ref{fig:sepfunction}, i.e., 	
\begin{equation}
	p_{t,k} = \dfrac{1}{1 + e^{-\vec{x}_{t}^{T}\vec{n}_{t,k}}}
\end{equation}
where the offset $c_{t,k} $ is included in the norm vector $\vec{n}_{t,k}$ as a bias term. In Fig. \ref{fig:sepfunction}, logistic regression functions for 1-dimensional case are shown for different parameters. Following the partitioning process, the second essential problem is to find out the linear models in each region. We assign a linear regressor specific to each distinct region and generate a corresponding estimate $\hat{y}_{t,r}$, given by 
\begin{equation}
	\hat{y}_{t,r} = \vec{w}_{t,r}^{T}\vec{x}_{t}
\end{equation}
where $\vec{w}_{t,r}$ is the regression vector particular to region $r$.	In the following subsections, we present different methods to partition the regressor space to construct our algorithms.
\subsection{Partitioning Methods}\label{subsec:parmethod}
We introduce two different partitioning methods: \textit{Type 1}, which is a straightforward partitioning and \textit{Type 2}, which is an efficient tree structured partitioning. 

\subsubsection{Type 1 Partitioning}\label{subsec:straightPar}

In this method, we allow each hyperplane to divide the whole space into two subspaces as shown in Fig. \ref{fig:straightpartition}. In order to clarify the technique, we work on the 2-dimensional space, i.e., the coordinate plane. Suppose, the observed feature vectors $\vec{x}_{t} = [x_{t,1},x_{t,2}]^{T} $ come from a bounded set $\{\Omega\}$ such that $-A \leq x_{t,1},x_{t,2} \leq A$ for some $A > 0$, as shown in Fig. \ref{fig:straightpartition}. We define 1-dimensional hyperplanes, whose normal vector representation is given by $\vec{n}_{t,k} \in \mathbbm{R}^{2}$ where $k$ denotes the corresponding region identity. At first, we have the whole space as a single set $\{\Omega\}$. Then we use a single separation function, which is a line in this case, to partition this space into subspaces $\{0\}$ and $\{1\}$ such that $\{0\} \cup \{1\} = \{\Omega\}$. When we add another hyperplane separating the set ${\Omega}$, we get four distinct subspaces $\{00\},\{01\},\{10\}$ and $\{11\}$ where their union forms the initial regression space. The number of separated regions increases by $O(k^{2})$. Note that if we use $k$ different separation functions, then we can obtain up to $\frac{k^{2}+k+2}{2}$ distinct regions forming a complete space. \par

\subsubsection{Type 2 Partitioning}\label{subsec:treePar}
\begin{figure}[t]
	\centering
	\includegraphics[scale=0.4]{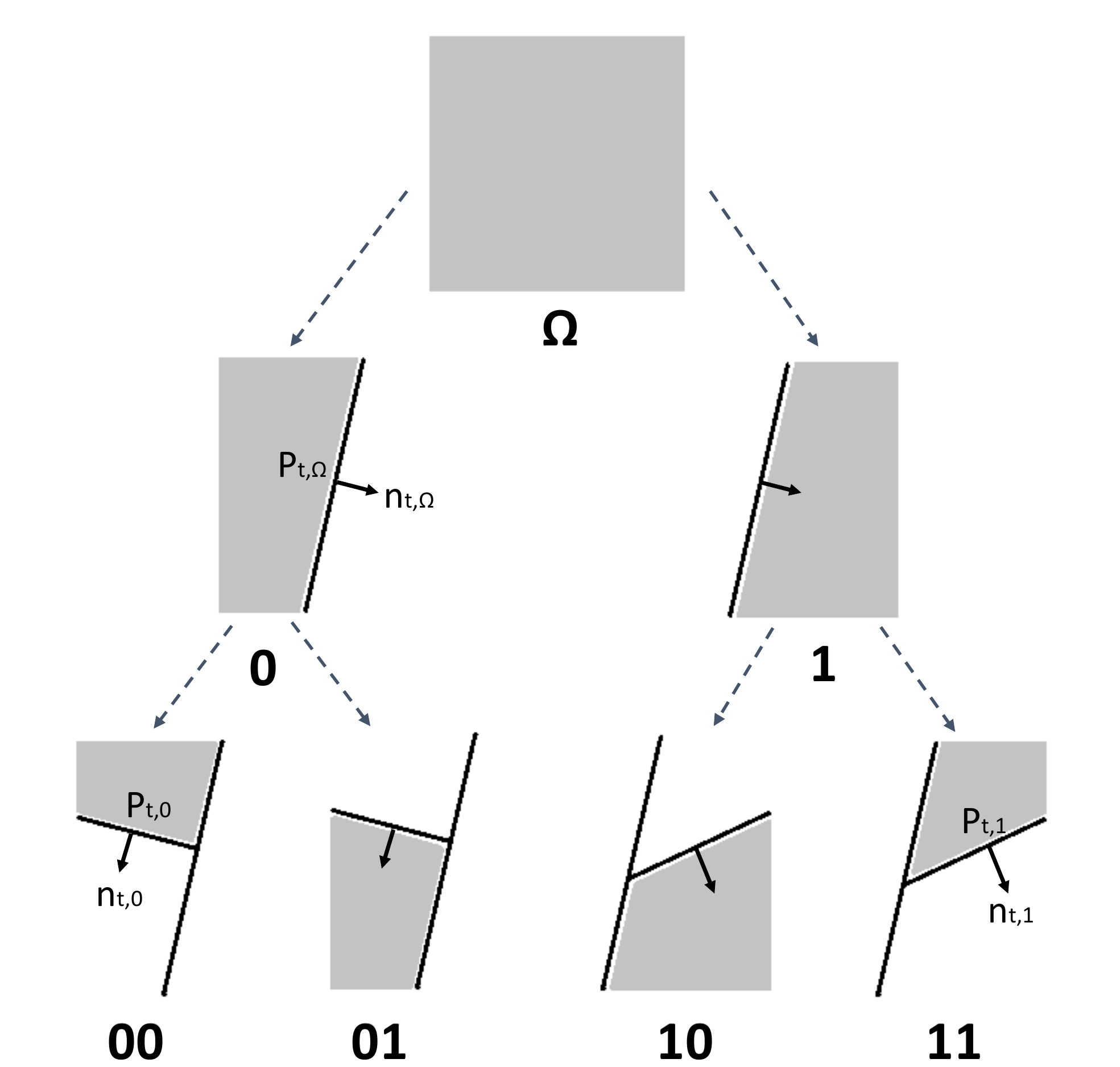} 
	\caption{Tree Based Partitioning of The Regression Space\label{fig:treepartition}}
\end{figure}

In the second method, we use the tree notion to partition the regression space, which is a more systematic way to determine the regions \cite{Hero,RPTrees}. We illustrate this method in Fig. \ref{fig:treepartition} for 2-dimensional case. First step is the same as previously mentioned approach, i.e., we partition the whole regression space into two distinct regions using one separation function. In the following steps, the partition technique is quite different. Since we have two distinct subspaces after the first step, we work on them separately, i.e., the partition process continues recursively in each subspace independent of the others. Therefore, adding one more hyperplane has an effect on just a single region, not on the whole space. The number of distinct regions in total increases by 1, when we apply one more separation function. Thus, in order to represent $p+1$ distinct regions, we specify $p$ separation functions. For the tree case, we use another identifier called the depth, which determines how deep the partition is, e.g. depth of the model shown in Fig. \ref{fig:treepartition} is 2. In particular, the number of different regions generated by the depth-$d$ models are given by $2^{d}$. Hence, the number of distinct regions increases in the order of $O(2^{d})$. For the tree based partitioning, we use the finest model of a depth-$d$ tree. The finest partition consists of the regions that are generated at the deepest level, e.g. regions $\{00\},\{01\},\{10\}$ and $\{11\}$ as shown in Fig. \ref{fig:treepartition}. \par
Both Type 1 and Type 2 partitioning have their own advantages, i.e., Type 2 partitioning achieves a better steady state error performance since the models generated by Type 1 partitioning are the subclasses of Type 2, however, Type 1 might perform better in the transient region since it uses less parameters.  
\subsection{Algorithm for Type 1 Partitioning}\label{subsec:algSB}
In this part, we introduce our first algorithm, which is based on the \textit{Type 1} partitioning. Following the model given in Fig. \ref{fig:straightpartition}, say, we have two different separator functions, $p_{t,0}, p_{t,1} \in \mathbbm{R}$, which are defined by $\vec{n}_{t,0},\vec{n}_{t,1} \in \mathbbm{R}^{2}$ respectively. For the region $\{00\}$, the corresponding estimate is given by
\[ \hat{y}_{t,00} = \vec{w}_{t,00}^{T}\vec{x}_{t}, \]
where $\vec{w}_{t,00} \in \mathbbm{R}^{2}$ is the regression vector of the region $\{00\}$. Since we have the estimates of all regions, the final estimate is given by
\begin{multline}
	\hat{y}_{t} = p_{t,0}p_{t,1}\hat{y}_{t,00} + p_{t,0}(1-p_{t,1})\hat{y}_{t,01} \\ +(1-p_{t,0})p_{t,1}\hat{y}_{t,10} + (1-p_{t,0})(1-p_{t,1})\hat{y}_{t,11}
\end{multline} 
when we observe the feature vector $\vec{x}_{t}$. This result can be easily extended to the cases where we have more then 2 separator functions.\par
We adaptively update the weights associated with each partition based on the overall performance. Boundaries of the regions are also updated to reach the best partitioning. We use the second order algorithms, e.g. Online Newton Step \cite{Hazan}, to update both separator functions and region weights. To accomplish this, the weight vector assigned to the region $\{00\}$ is updated as
\begin{equation}\label{w_t update}
	\begin{split}
		\vec{w}_{t+1,00} &= \vec{w}_{t,00} - \frac{1}{\beta}\vec{A}_{t}^{-1}\nabla e_{t}^{2} \\
		&= \vec{w}_{t,00} + \frac{2}{\beta}e_{t}p_{t,0}p_{t,1}\vec{A}_{t}^{-1}\vec{x}_{t},
	\end{split}
\end{equation} 
where $\beta$ is the step size, $\nabla$ is the gradient operator w.r.t. $\vec{w}_{t,00}$ and $\vec{A}_{t}$ is an $m\times m$ matrix defined as
\begin{equation}\label{A_t}
	\vec{A}_{t} = \sum_{i = 1}^{t}\nabla_{i} \nabla_{i}^{T} + \epsilon \vec{I}_{m},
\end{equation}
where $\nabla_t \triangleq \nabla e^2_t$ and $\epsilon > 0$ is used to ensure that $\vec{A}_{t}$ is positive definite, i.e., $\vec{A}_{t} > 0$, and invertible. Here, the matrix $\vec{A}_{t}$ is related to the Hessian of the error function, implying that the update rule uses the second order information \cite{Hazan}. \par
Region boundaries are also updated in the same manner. For example, the direction vector specifying the separation function $p_{t,0}$ in Fig. \ref{fig:straightpartition}, is updated as 
\begin{equation}\label{n_t update}
	\begin{split}
		\vec{n}_{t+1,0} &= \vec{n}_{t,0} - \frac{1}{\eta}\vec{A}_{t}^{-1}\nabla e_{t}^{2} \\
		&= \vec{n}_{t,0} + \frac{2}{\eta}e_{t}[p_{t,1}\hat{y}_{t,00} + (1 - p_{t,1})\hat{y}_{t,01} \\
		&- p_{t,1}\hat{y}_{t,10} - (1 - p_{t,1})\hat{y}_{t,11}]\vec{A}_{t}^{-1}\frac{\partial p_{t,0}}{\partial \vec{n}_{t,0}},
	\end{split}
\end{equation}
where $\eta$ is the step size to be determined, $\nabla$ is the gradient operator w.r.t. $\vec{n}_{t,0}$ and $\vec{A}_{t}$ is given in (\ref{A_t}). Partial derivative of the separation function $p_{t,0}$ w.r.t. $\vec{n}_{t,0}$ is given by 
\begin{equation}
	\dfrac{\partial p_{t,0}}{\partial \vec{n}_{t,0}} = \dfrac{\vec{x}_{t}e^{-\vec{x}_{t}^{T}\vec{n}_{t,0}}}{(1 + e^{-\vec{x}_{t}^{T}\vec{n}_{t,0}})^{2}}.
\end{equation}
All separation functions are updated in the same manner. In general, we derive the final estimate in a compact form as
\begin{equation}
	\hat{y}_{t} = \sum\limits_{r \in R}\hat{\psi}_{t,r},
\end{equation}
where $\hat{\psi}_{t,r}$ is the weighted estimate of region $r$ and $R$ represents the set of all region labels, e.g. $R = \{00, 01, 10, 11\}$ for the case given in Fig. \ref{fig:straightpartition}. Weighted estimate of each region is determined by
\begin{equation}
	\hat{\psi}_{t,r} = \hat{y}_{t,r} \prod\limits_{i = 1}^{K}\hat{p}_{t,P(i)},
\end{equation} 
where $K$ is the number of separation functions, $P$ represents the set of all separation function labels and $P(i)$ is the $i^{th}$ element of set $P$, e.g. $P = \{0, 1\}, P(1) = 0$, and $\hat{p}_{t,P(i)}$ is defined as
\begin{equation}
	\hat{p}_{t,P(i)} = \begin{cases} 
		p_{t,P(i)} &, r(i) = 0 \\
		1 - p_{t,P(i)} &, r(i) = 1 
	\end{cases},
\end{equation}
where $r(i)$ denotes the $i^{th}$ binary character of label $r$, e.g. $r = 10$ and $r(1) = 1$. We reformulate the update rules defined in (\ref{w_t update}) and (\ref{n_t update}) and present generic expressions for both regression weights and region boundaries. The derivations of the generic update rules are calculated after some basic algebra. Hence, the regression weights are updated as
\begin{equation} \label{w_t generic}
	\begin{split}
		\vec{w}_{t+1,r} = \vec{w}_{t,r} + \frac{2}{\beta}e_{t} \vec{A}_{t}^{-1} \vec{x}_{t} \prod\limits_{i=1}^{K} \hat{p}_{t,P(i)}
	\end{split}
\end{equation}
and the region boundaries are updated as
\begin{equation} \label{n_t generic}
	\begin{split}
		\vec{n}_{t+1,k} = \vec{n}_{t,k} + \frac{2}{\eta}e_{t} \vec{A}_{t}^{-1} \bigg[\sum\limits_{r \in R} \hat{y}_{t,r} (-1)^{r(i)} \prod\limits_{\substack{j = 1 \\ j \neq i}}^{K} \hat{p}_{t,P(j)} \bigg] \dfrac{\vec{x}_{t}e^{-\vec{x}_{t}^{T}\vec{n}_{t,k}}}{(1 + e^{-\vec{x}_{t}^{T}\vec{n}_{t,k}})^{2}},
	\end{split}
\end{equation}

\begin{algorithm}[t]
\caption{Straight Partitioning}
\begin{multicols}{2}
	\begin{algorithmic}[1]
		\State $\vec{A}_{0}^{-1} = \dfrac{1}{\epsilon} \vec{I}_{m}$
		\For{$t \leftarrow 1, n$}
		\State $\hat{y}_{t} \leftarrow 0$
		\ForAll{$r \in  R$}
		\State $\hat{y}_{t,r} \leftarrow \vec{w}_{t,r}^{T}\vec{x}_{t} $
		\State $\hat{\psi}_{t,r} \leftarrow \hat{y}_{t,r}$
		\State $\nabla_{t,r} \leftarrow \vec{x}_{t}$
		\For{$i \leftarrow 1, K$}
		\If{$r(i) := 0$}
		\State $\hat{p}_{t,P(i)} \leftarrow p_{t,P(i)}$
		\Else
		\State $\hat{p}_{t,P(i)} \leftarrow 1 - p_{t,P(i)}$
		\EndIf
		\State $\hat{\psi}_{t,r} \leftarrow \hat{\psi}_{t,r} \hat{p}_{t,P(i)}$
		\State $\nabla_{t,r} \leftarrow \nabla_{t,r}\hat{p}_{t,P(i)}$
		\EndFor
		\For{$i \leftarrow 1, K$}
		\State $\alpha_{t,P(i)} \leftarrow (-1)^{r(i)}(\hat{\psi}_{t,r} / \hat{p}_{t,P(i)})$
		\EndFor
		\State $\hat{y}_{t} \leftarrow \hat{y}_{t} + \hat{\psi}_{t,r} $
		\EndFor
		\State $e_{t} \leftarrow y_{t} - \hat{y}_{t}$
		
		\ForAll{$r \in  R$}
		\State $\nabla_{t,r} \leftarrow -2e_{t}\nabla_{t,r}$
		\State $\vec{A}_{t,r}^{-1} \leftarrow \vec{A}_{t-1,r}^{-1} - \dfrac{\vec{A}_{t-1,r}^{-1} \nabla_{t,r} \nabla_{t,r}^{T} \vec{A}_{t-1,r}^{-1}}{1 + \nabla_{t,r}^{T} \vec{A}_{t-1,r}^{-1} \nabla_{t,r}}$
		\State $\vec{w}_{t+1,r} \leftarrow \vec{w}_{t,r} - \dfrac{1}{\beta}\vec{A}_{t,r}^{-1}\nabla_{t,r}$
		\EndFor
		\For{$i \leftarrow 1, K$}
		\State$k \leftarrow P(i)$
		\State$\nabla_{t,k} \leftarrow -2e_{t} \alpha_{t,k} p_{t,k}(1 - p_{t,k})\vec{x_{t}}$
		\State$\vec{A}_{t,k}^{-1} \leftarrow \vec{A}_{t-1,k}^{-1} - \dfrac{\vec{A}_{t-1,k}^{-1} \nabla_{t,k} \nabla_{t,k}^{T} \vec{A}_{t-1,k}^{-1}}{1 + \nabla_{t,k}^{T} \vec{A}_{t-1,k}^{-1} \nabla_{t,k}}$
		\State$\vec{n}_{t+1,k} \leftarrow \vec{n}_{t,k} - \dfrac{1}{\eta} \vec{A}_{t,k}^{-1}\nabla_{t,k}$
		\EndFor
		\EndFor
	\end{algorithmic}
\end{multicols}
\end{algorithm}
where we assign $k = P(i)$, i.e., separation function with label-$k$ is the $i^{th}$ entry of set $P$. 
Partial derivative of the logistic regression function $p_{t,k}$ w.r.t. $\vec{n}_{t,k}$ is also inserted in (12).
In order to avoid taking the inverse of an $m \times m$ matrix, $\vec{A}_{t}$, at each iteration in (\ref{w_t generic}) and (\ref{n_t generic}), we generate a recursive formula using matrix inversion lemma for $\vec{A}_{t}^{-1}$ given as \cite{sayedbook}
\begin{equation}
	\vec{A}_{t}^{-1} = \vec{A}_{t-1}^{-1} - \dfrac{\vec{A}_{t-1}^{-1} \nabla_{t} \nabla_{t}^{T} \vec{A}_{t-1}^{-1}}{1 + \nabla_{t}^{T} \vec{A}_{t-1}^{-1} \nabla_{t}},
\end{equation}  
where $\nabla_{t} \triangleq \nabla e_{t}^{2}$ w.r.t. the corresponding variable. The complete algorithm for \textit{Type 1} partitioning is given in Algorithm 1 with all updates and initializations.

\subsection{Algorithm for Type 2 Partitioning}\label{subsec:algFM}
In this algorithm, we use another approach to estimate the desired data. The partition of the regressor space will be based on the finest model of a tree structure \cite{Hero,Helmbold}. We follow the case given in Fig. \ref{fig:treepartition}. Here, we have three separation functions, $p_{t,\varepsilon}, p_{t,0}$ and $p_{t,1}$, partitioning the whole space into four subspaces. The corresponding direction vectors are given by $\vec{n}_{t,\varepsilon}, \vec{n}_{t,0}$ and $\vec{n}_{t,1}$ respectively. Using the individual estimates of all four regions, we find the final estimate by
\begin{multline}
\hat{y}_{t} = p_{t,\varepsilon}p_{t,0}\hat{y}_{t,00} + p_{t,\varepsilon}(1-p_{t,0})\hat{y}_{t,01} \\ +(1-p_{t,\varepsilon})p_{t,1}\hat{y}_{t,10} + (1-p_{t,\varepsilon})(1-p_{t,1})\hat{y}_{t,11}
\end{multline} 
which can be extended to depth-$d$ models with $d > 2$.\par
\begin{figure}[t]
	\centering
	\includegraphics[scale=0.35]{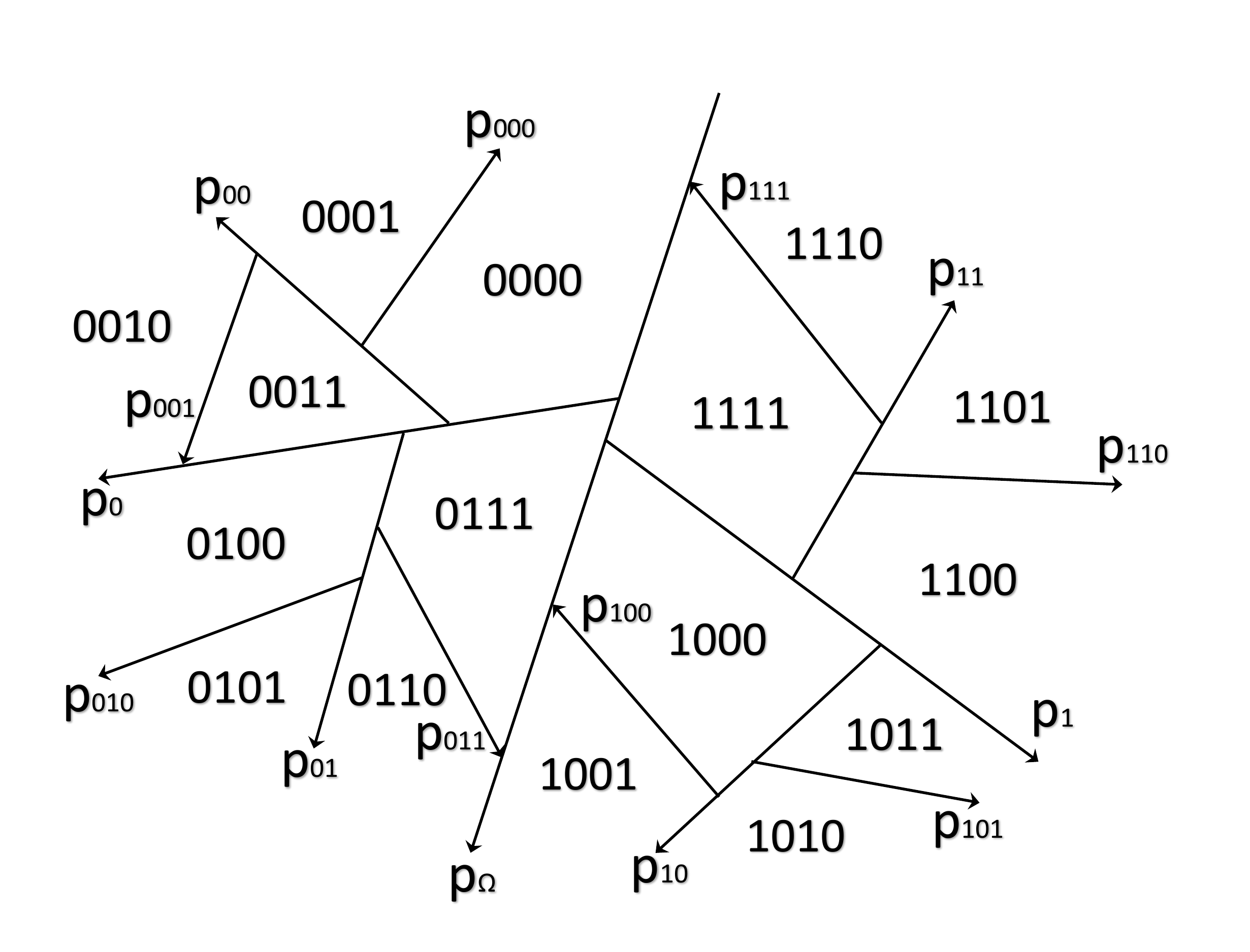} 
	\caption{Labeling Example for the Depth-4 Case of the Finest Model  \label{fig:finlabel}}
\end{figure}

%
Regressors of each region is updated similar to the first algorithm. We demonstrate a systematic way of labeling for partitions in Fig. \ref{fig:finlabel}. The final estimate of this algorithm is given by the following generic formula
\begin{equation}
\hat{y}_{t} = \sum\limits_{j = 1}^{2^{d}}\hat{\psi}_{t,R_{d}(j)}
\end{equation}
where $R_{d}$ is the set of all region labels with length $d$ in the increasing order for, i.e., $R_{1} = \{0, 1\}$ or $R_{2} = \{00, 01, 10, 11\}$ and $R_{d}(j)$ represents the $j^{th}$ entry of set $R_{d}$. Weighted estimate of each region is found as 
\begin{equation}
\hat{\psi}_{t,r} = \hat{y}_{t,r} \prod\limits_{i = 1}^{d}\hat{p}_{t,r_{i}}
\end{equation}
where $r_{i}$ denotes the first $i-1$ character of label $r$ as a string, i.e., $r = \{0101\}, r_{3} = \{01\}$ and $r_{1} = \{\epsilon\}$, which is the empty string $\{\epsilon\}$. Here, $\hat{p}_{t,r_{i}}$ is defined as
\begin{equation}
\hat{p}_{t,r_{i}} = \begin{cases} 
p_{t,r_{i}} &, r(i) = 0 \\
1 - p_{t,r_{i}} &, r(i) = 1 
\end{cases}.
\end{equation}\par
Update rules for the region weights and the boundaries are given as a generic form and the derivations of these updates are obtained after some basic algebra. Regressor vectors are updated as 
\begin{equation}
\vec{w}_{t+1,r} = \vec{w}_{t,r} + \dfrac{2}{\beta}e_{t} \vec{A}_{t}\vec{x}_{t} \prod\limits_{i = 1}^{d}\hat{p}_{t,r_{i}}  
\end{equation}
\begin{algorithm}[t]
	\caption{Finest Model Partitioning}
\begin{multicols}{2}
	\begin{algorithmic}[1]
		\State $A_{0}^{-1} \leftarrow \dfrac{1}{\epsilon} \vec{I}_{m}$
		\For{$t \leftarrow 1, n$}
		\State $\hat{y}_{t} \leftarrow 0$
		\For{$j \leftarrow 1, 2^{d}$}
		\State $r \leftarrow R_{d}(j)$
		\State $\hat{y}_{t,r} \leftarrow \vec{w}_{t,r}^{T}\vec{x}_{t} $
		\State $\hat{\psi}_{t,r} \leftarrow \hat{y}_{t,r}$
		\State $\gamma_{t,r} \leftarrow 1$
		\For{$i \leftarrow 1, d$}
		\If{$r(i) \leftarrow 0$}
		\State $\hat{p}_{t,r_{i}}\leftarrow p_{t,r_{i}}$
		\Else		
		\State$\hat{p}_{t,r_{i}}\leftarrow 1 - p_{t,r_{i}}$
		\EndIf
		\State $\hat{\psi}_{t,r} 
		\leftarrow \hat{\psi}_{t,r}\hat{p}_{t,r_{i}}$
		\State $\gamma_{t,r} \leftarrow \gamma_{t,r}\hat{p}_{t,r_{i}}$
		\EndFor
		\State $\hat{y}_{t} \leftarrow \hat{y}_{t} + \hat{\psi}_{t,r} $
		\EndFor
		\For{$i \leftarrow 1, 2^{d}-1$}
		\State$k \leftarrow P(i)$
		\For{$j \leftarrow 1, 2^{d-\ell(k)}$}
		\State$r \leftarrow concat[k:R_{d-\ell(k)}(j)]$
		\State$\alpha_{t,k} \leftarrow (-1)^{r(\ell(k)+1)}(\hat{\psi}_{t,r}/ \hat{p}_{t,k})$
		\EndFor
		\EndFor
		\State $e_{t} \leftarrow y_{t} - \hat{y}_{t}$
		\For{$j \leftarrow 1, 2^{d}$}
		\State $r \leftarrow R_{d}(j)$
		\State $\nabla_{t,r} \leftarrow -2e_{t} \gamma_{t,r} \vec{x_{t}}$
		\State $\vec{A}_{t,r}^{-1} \leftarrow \vec{A}_{t-1,r}^{-1} - \dfrac{\vec{A}_{t-1}^{-1} \nabla_{t,r} \nabla_{t,r}^{T} \vec{A}_{t-1,r}^{-1}}{1 + \nabla_{t,r}^{T} \vec{A}_{t-1,r}^{-1} \nabla_{t,r}}$
		\State $\vec{w}_{t+1,r} \leftarrow \vec{w}_{t,r} - \dfrac{1}{\beta}\vec{A}_{t,r}^{-1}\nabla_{t,r} $
		\EndFor
		\For{$i \leftarrow 1, 2^{d}-1$}
		\State $k \leftarrow P(i)$
		\State $\nabla_{t,k} \leftarrow -2e_{t} \alpha_{t,k} p_{t,k}(1 - p_{t,k})\vec{x_{t}}$
		\State $\vec{A}_{t,k}^{-1} \leftarrow \vec{A}_{t-1,k}^{-1} - \dfrac{\vec{A}_{t-1,k}^{-1} \nabla_{t,k} \nabla_{t,k}^{T} \vec{A}_{t-1,k}^{-1}}{1 + \nabla_{t,k}^{T} \vec{A}_{t-1,k}^{-1} \nabla_{t,k}}$
		\State $\vec{n}_{t+1,k} \leftarrow \vec{n}_{t,k} - \dfrac{1}{\eta}\vec{A}_{t,k}^{-1}\nabla_{t,k} $
		\EndFor
		\EndFor
	\end{algorithmic}
\end{multicols}
\end{algorithm}
and the separator function updates are given by
\begin{equation}
\begin{split}
\vec{n}_{t+1,k} = \vec{n}_{t,k}
+ \frac{2}{\eta}e_{t} \vec{A}_{t}^{-1} \bigg[\sum\limits_{j = 1}^{2^{d-\ell(k)}} \hat{y}_{t,r} (-1)^{r(\ell(k) + 1)} \prod\limits_{\substack{i = 1 \\ r_{i} \neq k}}^{d} \hat{p}_{t,r_{i}} \bigg] \dfrac{\partial p_{t,k}}{\partial \vec{n}_{t,k}}
\end{split}
\end{equation}	
where $r$ is the label string generated by concatenating separation function id $k$ and the label kept in $j^{th}$ entry of the set $R_{(d-\ell(k))}$, i.e., $r = [k;R_{(d-\ell(k))}(j)]$ and $\ell(k)$ represents the length of binary string $k$, e.g. $\ell(01) = 2$. The partial derivative of $p_{t,k}$ w.r.t. $\vec{n}_{t,k}$ is the same expression given in (14). The complete algorithm for \textit{Type 2} partitioning is given in Algorithm 2 with all updates and initializations. 
%
%
\subsection{Algorithm for Combining All Possible Models of Tree}\label{subsec:algComb}
\begin{figure}
	\centering
	\includegraphics[scale=0.4]{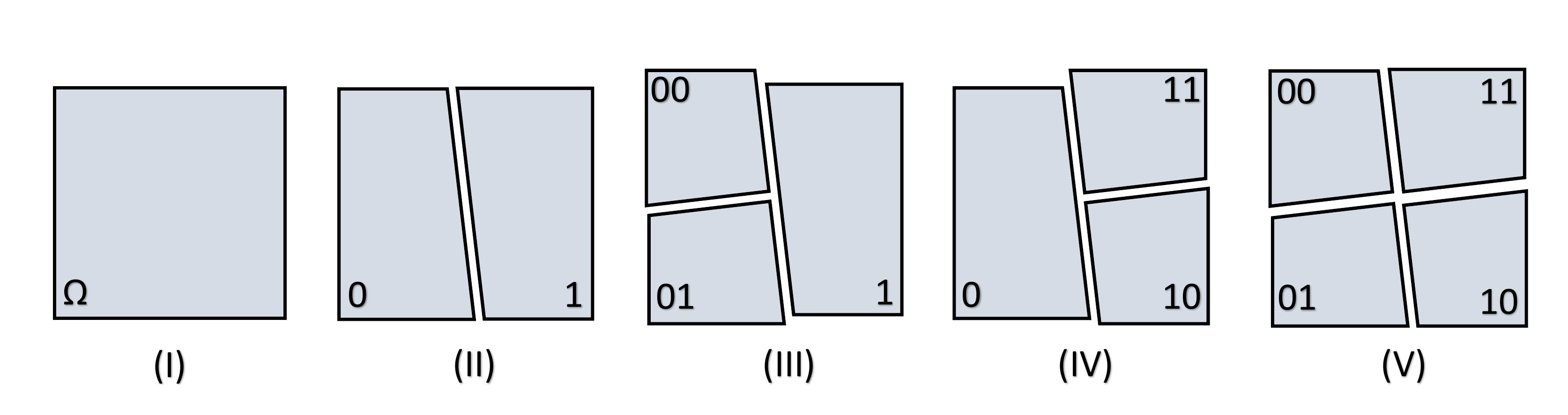} 
	\caption{All Possible Models for the Depth-2 Tree Based Partitioning  \label{fig:treeModels}}
\end{figure}

In this algorithm, we combine the estimates generated by all possible models of a tree based partition, instead of considering only the finest model. The main goal of this algorithm is to illustrate that using only the finest model of a depth-$d$ tree provides a better performance. For example, we represent the possible models corresponding to a depth-2 tree in Fig. \ref{fig:treeModels}. We emphasize that the last partition is the finest model we use in the previous algorithm. Following the case in Fig. \ref{fig:treeModels}, we generate five distinct piecewise linear models and estimates of these models. The final estimate is then constructed by linearly combining the outputs of each piecewise linear model, represented by $\hat{\phi}_{t,\lambda}$, where $\lambda$ represents the model identity. Hence, $\hat{y}_{t}$ is given by 
\begin{equation}
\hat{y}_{t} = \vec{\upsilon}_{t}^{T}\hat{\vec{\phi}}_{t}
\end{equation}
where  $\hat{\vec{\phi}}_{t} = [\hat{\phi}_{t,1}, \hat{\phi}_{t,2}, ..., \hat{\phi}_{t,M}]^{T}$, $\vec{\upsilon}_{t} \in \mathbbm{R}^{M}$ is the weight vector and $M$ represents the number of possible distinct models generated by a depth-$d$ tree, e.g. $M = 5$ for depth-2 case. In general, we have $M \approx (1.5)^{2^{d}}$. Model estimates, $\hat{\phi}_{t,\lambda}$, are calculated in the same way as in Section \ref{subsec:algFM}. Linear combination weights, $\vec{v}_t$, are also adaptively updated using the second order methods as performed in the previous sections.
 
\subsection{Computational Complexities}\label{subsec:compComplexity}
In this section, we determine the computational complexities of the proposed algorithms. In the algorithm for \textit{Type 1} partitioning, the regressor space is partitioned into at most $\frac{k^2 + k + 2}{2}$ regions by using $k$ distinct separator function. Thus, this algorithm requires $O(k^2)$ weight update at each iteration. In the algorithm for \textit{Type 2} partitioning, the regressor space is partitioned into $2^{d}$ regions for the depth-$d$ tree model. Hence, we perform $O(2^{d})$ weight update at each iteration. The last algorithm combines all possible models of depth-$d$ tree and calculates the final estimate in an efficient way requiring $O(4^d)$ weight updates \cite{DAT}. Suppose that the regressor space is $m$-dimensional, i.e., $\vec{x}_{t} \in \mathbbm{R}^{m}$. For each update, all three algorithms require $O(m^2)$ multiplication and addition resulting form a matrix-vector product, since we apply second order update methods. Therefore, the corresponding complexities are $O(m^{2}k^{2})$, $O(m^{2}2^{d})$ and $O(m^{2}4^{d})$ for the Algorithm 1, the Algorithm 2 and the Algorithm 3 respectively. In Table \ref{table:Complexity}, we represent the computational complexities of the existing algorithms. "FMP" and "SP" represents Finest Model Partitioning and Straight Partitioning algorithms respectively. "DFT" stands for Decision Fixed Tree and "DAT" represents Decision Adaptive Tree \cite{DAT}. "S-DAT" denotes the Decision Adaptive Tree with second order update rules. "CTW" is used for Context Tree Weighting \cite{CTW}, "GKR" represents Gaussian-Kernel regressor \cite{kernel}, "VF" represents Volterra Filter \cite{volterra}, "FNF" and "EMFNF" stand for the Fourier and Even Mirror Fourier Nonlinear Filter \cite{fnf} respectively.
\begin{table}[t]
	\centering
	\caption{Computational Complexities \label{table:Complexity}}
	\begin{tabularx}{12cm}{ |c||C|C|C|C|C| }
		\hline
		Algorithms & FMP & SP & S-DAT & DFT & DAT \\ \hline 
		Complexity & $O(m^{2}2^{d})$ & $O(m^{2}k^{2})$ & $O(m^{2}4^{d})$ & $O(md2^{d})$ & $O(m4^{d})$ \\ \hline \hline
		Algorithms & GKR & CTW & FNF & EMFNF & VF \\ \hline 
		Complexity & $O(m2^{d})$ & $O(md)$ & $O(m^{n}n^{n})$ & $O(m^{n})$ & $O(m^{n})$ \\ \hline  
	\end{tabularx}
\end{table} 

\subsection{Logarithmic Regret Bound}\label{subsec:regretLOG}
In this subsection, we provide regret results for the introduced algorithms. All three algorithms uses the second order update rule, Online Newton Step \cite{Hazan}, and achieves a logarithmic regret when the normal vectors of the region boundaries are fixed and the cost function is convex in the sense of individual region weights. In order to construct the upper bounds, we first let $\vec{w}_{n}^{*}$ be the best predictor in hindsight, i.e.,
\begin{equation}
\vec{w}_{n}^{*} = \argmin_{\vec{w}}\sum\limits_{t=1}^{n}e_{t}^{2}(\vec{w})
\end{equation}
and express the following inequality
\begin{equation}
\begin{split}
e_{t}^{2}(\vec{w}_{t}) - e_{t}^{2}(\vec{w}_{n}^{*}) \leq \nabla_{t}^{T}(\vec{w}_{t}-\vec{w}_{n}^{*}) 
- \dfrac{\beta}{2}(\vec{w}_{t}-\vec{w}_{n}^{*})^{T} \nabla_{t}\nabla_{t}^{T} (\vec{w}_{t}-\vec{w}_{n}^{*})
\end{split}
\end{equation}
using the Lemma 3 of \cite{Hazan}, since our cost function is $\alpha$-exp-concave, i.e., exp$(-\alpha e_{t}^{2}(\vec{w}_{t}))$ is concave for $\alpha > 0$ and has an upper bound $G$ on its gradient, i.e., $\|\nabla_{t}\| \leq G$. We give the update rule for regressor weights as 
\begin{equation}
\vec{w}_{t+1} = \vec{w}_{t} - \dfrac{1}{\beta}\vec{A}_{t}^{-1}\nabla_{t}.
\end{equation}
When we subtract the optimal weight from both sides, we get
\begin{align}
&\vec{w}_{t+1} - \vec{w}_{n}^{*} = \vec{w}_{t} - \vec{w}_{n}^{*} - \dfrac{1}{\beta}\vec{A}_{t}^{-1}\nabla_{t} \\
&\vec{A}_{t}(\vec{w}_{t+1} - \vec{w}_{n}^{*}) = \vec{A}_{t}(\vec{w}_{t} - \vec{w}_{n}^{*}) - \dfrac{1}{\beta}\nabla_{t}
\end{align}
and multiply second equation with the transpose of the first equation to get
\begin{equation} \label{26}
\begin{split}
\nabla_{t}(\vec{w}_{t} - \vec{w}_{n}^{*}) &= \dfrac{1}{2\beta}\nabla_{t}^{T}\vec{A}_{t}^{-1}\nabla_{t} + \dfrac{\beta}{2}(\vec{w}_{t} - \vec{w}_{n}^{*})^{T}\vec{A}_{t}(\vec{w}_{t} - \vec{w}_{n}^{*}) \\ 
&- \dfrac{\beta}{2}(\vec{w}_{t+1} - \vec{w}_{n}^{*})^{T}\vec{A}_{t}(\vec{w}_{t+1} - \vec{w}_{n}^{*}). 
\end{split}
\end{equation}
By following a similar discussion \cite{Hazan}, except that we have equality in (\ref{26}) and in the proceeding parts, we achieve the inequality 
\begin{equation}
\begin{split}
\sum\limits_{t = 1}^{n}S_{t} \leq \dfrac{1}{2\beta}\sum\limits_{t = 1}^{n} \nabla_{t}^{T}\vec{A}_{t}^{-1}\nabla_{t}  
+ \dfrac{\beta}{2}(\vec{w}_{1} - \vec{w}_{n}^{*})^{T}\vec{A}_{0}(\vec{w}_{1} - \vec{w}_{n}^{*})
\end{split},
\end{equation}
where $S_{t}$ is defined as
\begin{equation}
\begin{split}
S_{t} \triangleq \nabla_{t}^{T}(\vec{w}_{t}-\vec{w}_{n}^{*})- \dfrac{\beta}{2}(\vec{w}_{t}-\vec{w}_{n}^{*})^{T} \nabla_{t}\nabla_{t}^{T} (\vec{w}_{t}-\vec{w}_{n}^{*}).
\end{split}
\end{equation}
Since we define $\vec{A}_{0} = \epsilon\vec{I}_{m}$ and have a finite space of regression vectors, i.e., $\|\vec{w}_{t}-\vec{w}_{n}^{*} \|^{2} \leq A^{2}$, we get
\begin{equation}
\begin{split}
\sum\limits_{t=1}^{n}e_{t}^{2}(\vec{w}_{t}) - \sum\limits_{t=1}^{n}e_{t}^{2}(\vec{w}_{n}^{*}) &\leq \dfrac{1}{2\beta}\sum\limits_{t = 1}^{n} \nabla_{t}^{T}\vec{A}_{t}^{-1}\nabla_{t} + \dfrac{\beta}{2}\epsilon \delta^{2} \\
& \leq \dfrac{1}{2\beta}\sum\limits_{t = 1}^{n} \nabla_{t}^{T}\vec{A}_{t}^{-1}\nabla_{t} + \dfrac{1}{2\beta},
\end{split}
\end{equation}
where we choose $\epsilon = \frac{1}{\beta^{2}A^{2}}$ and use the inequalities (10) and (17). Now, we specify an upper bound for the first term in LHS of the inequality (19).  We make use of Lemma 11 given in \cite{Hazan}, to get the following bound
\begin{equation}
\begin{split}
\dfrac{1}{2\beta}\sum\limits_{t = 1}^{n} \nabla_{t}^{T}\vec{A}_{t}^{-1}\nabla_{t} \leq \dfrac{m}{2\beta}\log \bigg(\dfrac{G^{2}n}{\epsilon}+1\bigg) 
 = \dfrac{m}{2\beta}\log (G^{2}n\beta^{2}A^{2}+1) 
 \leq \dfrac{m}{2\beta}\log(n),
\end{split}
\end{equation}
where in the last inequality, we use the choice of $\beta$, i.e., $\beta = \frac{1}{2}\min\{\frac{1}{4GA},\alpha\}$, which implies that $\frac{1}{\beta} \leq 8(GA+\frac{1}{\alpha})$. Therefore, we present the final logarithmic regret bound as
\begin{equation}
\sum\limits_{t=1}^{n}e_{t}^{2}(\vec{w}_{t}) - \sum\limits_{t=1}^{n}e_{t}^{2}(\vec{w}_{n}^{*}) \leq 5\bigg(GA+\dfrac{1}{\alpha}\bigg)m\log(n).
\end{equation}

\section{Simulations}\label{sec:simulations}
In this section, we evaluate the performance of the proposed algorithms under different scenarios. In the first set of simulations, we aim to provide a better understanding of our algorithms. To this end, we first consider the regression of a signal that is generated by a  piecewise linear model whose partitions match the initial partitioning of our algorithms. Then we examine the case of mismatched initial partitions to illustrate the learning process of the presented algorithms. As the second set of simulation, we mainly assess the merits of our algorithms by using the well known real and synthetic benchmark datasets that are extensively used in the signal processing and the machine learning literatures, e.g., California Housing \cite{ltorgo}, Kinematics \cite{ltorgo} and Elevators \cite{ltorgo}. We then perform two more experiments with two chaotic processes, e.g., the Gauss map and the Lorenz attractor, to demonstrate the merits of our algorithms. All data sequences used in the simulations are scaled to the range $[-1,1]$ and the learning rates are selected to obtain the best steady state performance of each algorithm.\par      
\begin{figure}
	\centering
	\includegraphics[scale=0.7]{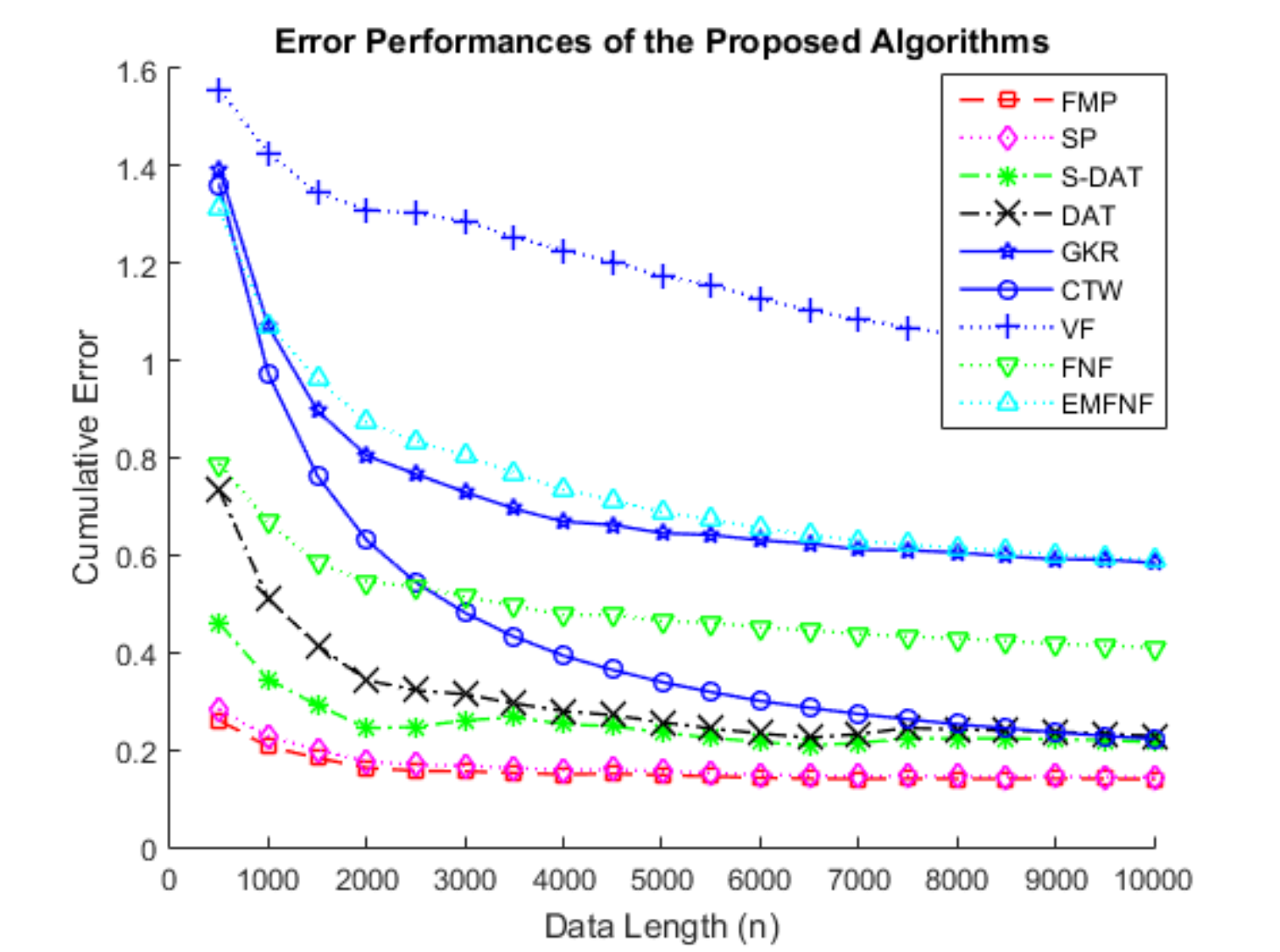} 
	\caption{Regression error performances for the matched partitioning case using model (\ref{matched_Par}).  \label{fig:matched}}
\end{figure}
\subsection{Matched Partition}\label{sec:matchedPar}
In this subsection, we consider the regression of a signal generated using a piecewise linear model whose partitions match with the initial partitioning of the proposed algorithms. The main goal of this experiment is to provide an insight on the working principles of the proposed algorithms. Hence, this experiment is not designated to assess the performance of our algorithms with respect to the ones that are not based on piecewise linear modeling. This is only an illustration of how it is possible to achieve a performance gain when the data sequence is generated by a nonlinear system. \par
We use the following piecewise linear model to generate the data sequence,
\begin{equation}\label{matched_Par}
\hat{y}_{t} = \begin{cases} 
\vec{w}_{1}^{T}\vec{x}_{t} + \upsilon_{t} &, \vec{x}_{t}^{T}\vec{n}_0 \geq 0 \text{ and }\vec{x}_{t}^{T}\vec{n}_1 \geq 0 \\
\vec{w}_{2}^{T}\vec{x}_{t} + \upsilon_{t} &, \vec{x}_{t}^{T}\vec{n}_0 \geq 0 \text{ and }\vec{x}_{t}^{T}\vec{n}_1 < 0 \\
\vec{w}_{2}^{T}\vec{x}_{t} + \upsilon_{t} &, \vec{x}_{t}^{T}\vec{n}_0 < 0 \text{ and }\vec{x}_{t}^{T}\vec{n}_1 \geq 0 \\
\vec{w}_{1}^{T}\vec{x}_{t} + \upsilon_{t} &, \vec{x}_{t}^{T}\vec{n}_0 < 0 \text{ and }\vec{x}_{t}^{T}\vec{n}_1 < 0 \\
\end{cases}
\end{equation}
where $\vec{w}_{1} = [1,1]^{T}$, $\vec{w}_{2} = [-1,-1]^{T}$, $\vec{n}_{0} = [1,0]^{T}$ and $\vec{n}_{1} = [0,1]^{T}$. The feature vector $\vec{x}_{t} = [x_{t,1}, x_{t,2}]^{T}$ is composed of two jointly Gaussian processes with $[0, 0]^{T}$ mean and $\vec{I}_{2}$ variance. $\upsilon_{t}$ is a sample taken from a Gaussian process with zero mean and 0.1 variance. The generated data sequence is represented by $\hat{y}_{t}$. In this scenario, we set the learning rates to 0.125 for the FMP, 0.0625 for the SP, 0.005 for the S-DAT, 0.01 for the DAT, 0.5 for the GKR, 0.004 for the CTW, 0.025 for the VF and the EMFNF, 0.005 for the FNF. \par
In Fig. \ref{fig:matched}, we represent the deterministic error performance of the specified algorithms. The algorithms VF, EMFNF, GKR and FNF cannot capture the characteristic of the data model, since these algorithms are constructed to achieve satisfactory results for smooth nonlinear models, but we examine a highly nonlinear and discontinuous model. On the other hand, the algorithms FMP, SP, S-DAT, CTW and DAT attain successive performance due to their capability of handling highly nonlinear models. As seen in Fig. \ref{fig:matched}, our algorithms, the FMP and the SP, significantly outperform their competitors and achieve almost the same performance result, since the data distribution is completely captured by both algorithms. Although the S-DAT algorithm does not perform as well as the FMP and the SP algorithms, still obtains a better convergence rate compared to the DAT and the CTW algorithms.

\subsection{Mismatched Partition}\label{sec:mismatchedPar}
In this subsection, we consider the case where the desired data is generated by a piecewise linear model whose partitions do not match with the initial partitioning of the proposed algorithms. This experiment mainly focuses on to demonstrate how the proposed algorithms learn the underlying data structure. We also aim to emphasize the importance of adaptive structure. \par
We use the following piecewise linear model to generate the data sequence,
\begin{equation} \label{mismatcedPar}
\hat{y}_{t} = \begin{cases} 
\vec{w}_{1}^{T}\vec{x}_{t} + \upsilon_{t} &, \vec{x}_{t}^{T}\vec{n}_0 \geq 0.5 \text{ and }\vec{x}_{t}^{T}\vec{n}_1 \geq -0.5 \\
\vec{w}_{2}^{T}\vec{x}_{t} + \upsilon_{t} &, \vec{x}_{t}^{T}\vec{n}_0 \geq 0.5 \text{ and }\vec{x}_{t}^{T}\vec{n}_1 < -0.5 \\
\vec{w}_{2}^{T}\vec{x}_{t} + \upsilon_{t} &, \vec{x}_{t}^{T}\vec{n}_0 < 0.5 \text{ and }\vec{x}_{t}^{T}\vec{n}_2 \geq -0.5 \\
\vec{w}_{1}^{T}\vec{x}_{t} + \upsilon_{t} &, \vec{x}_{t}^{T}\vec{n}_0 < 0.5 \text{ and }\vec{x}_{t}^{T}\vec{n}_2 < -0.5 \\
\end{cases}
\end{equation}
where $\vec{w}_{1} = [1,1]^{T}$, $\vec{w}_{2} = [1,-1]^{T}$, $\vec{n}_{0} = [2,-1]^{T}$, $\vec{n}_{1} = [-1,1]^{T}$ and $\vec{n}_{2} = [2,1]^{T}$. The feature vector $\vec{x}_{t} = [x_{t,1}, x_{t,2}]^{T}$ is composed of two jointly Gaussian processes with $[0, 0]^{T}$ mean and $\vec{I}_{2}$ variance. $\upsilon_{t}$ is a sample taken from a Gaussian process with zero mean and 0.1 variance. The generated data sequence is represented by $\hat{y}_{t}$. The learning rates are set to 0.04 for the FMP, 0.025 for the SP, 0.005 for the S-DAT, the CTW and the FNF, 0.025 for the EMFNF and the VF, 0.5 for the GKR. \par
\begin{figure}
	\centering
	\includegraphics[scale=0.7]{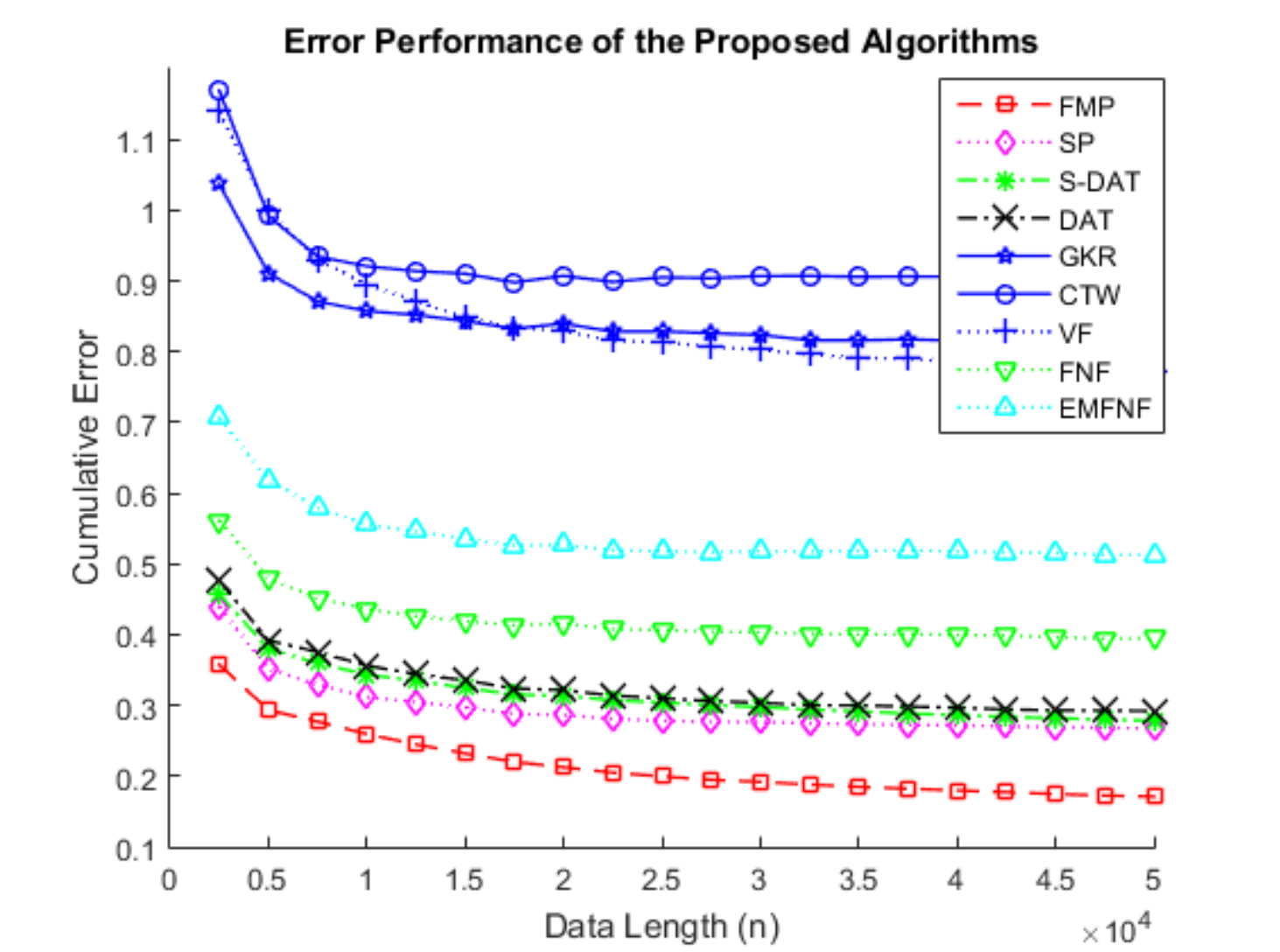} 
	\caption{Regression error performances for the mismatched partitioning case using model (\ref{mismatcedPar}).  \label{fig:mismatched}}
\end{figure}
In Fig. \ref{fig:mismatched}, we demonstrate the normalized time accumulated error performance of the proposed algorithms. Different from the matched partition scenario, we emphasize that the CTW algorithm performs even worse than the VF, the FNF and the EMFNF algorithms, which are not based on piecewise linear modeling. The reason is that the CTW algorithm has fixed regions that are mismatched with the underlying partitions. Besides, the adaptive algorithms, FMP, SP, S-DAT and DAT achieve considerably better performance, since these algorithms update their partitions in accordance with the data distribution. Comparing these four algorithms, Fig. \ref{fig:mismatched} exhibits that the FMP notably outperforms its competitors, since this algorithm exactly matches its partitioning to the partitions of the piecewise linear model given in (\ref{mismatcedPar}).\par
\begin{figure}
	\centering
	\includegraphics[scale=0.29]{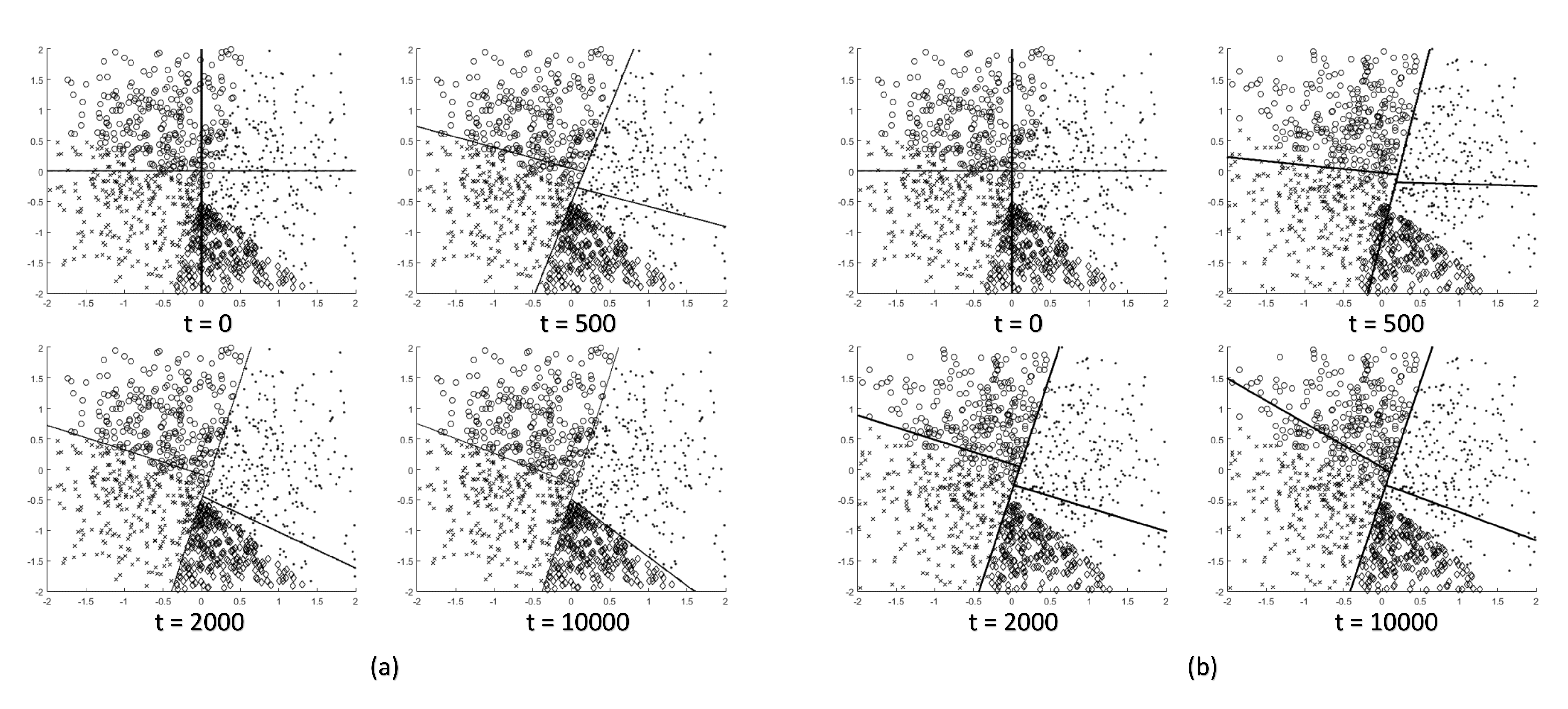} 
	\caption{Training of the separation functions for the mismatched partitioning scenario (a) FMP Algorithm (b) DAT Algorithm. \label{fig:rotatingSep}}
\end{figure}
We illustrate how the FMP and the DAT algorithms update their region boundaries in Fig. \ref{fig:rotatingSep}. Both algorithms initially partition the regression space into 4 equal quadrant, i.e., the cases shown in $t=0$. We emphasize that when the number of iterations reaches 10000, i.e., $t = 10000$, the FMP algorithm trains its region boundaries such that its partitions substantially match the  partitioning of the piecewise linear model. However, the DAT algorithm cannot capture the data distribution yet, when $t = 10000$. Therefore, the FMP algorithm, which uses the second order methods for training, has a faster convergence rate compared to the DAT algorithm, which updates its region boundaries using first order methods. 
\subsection{Real and Synthetic Data Sets}\label{sec:realData}  
\begin{figure}[t]
	\centering
	\includegraphics[scale=0.7]{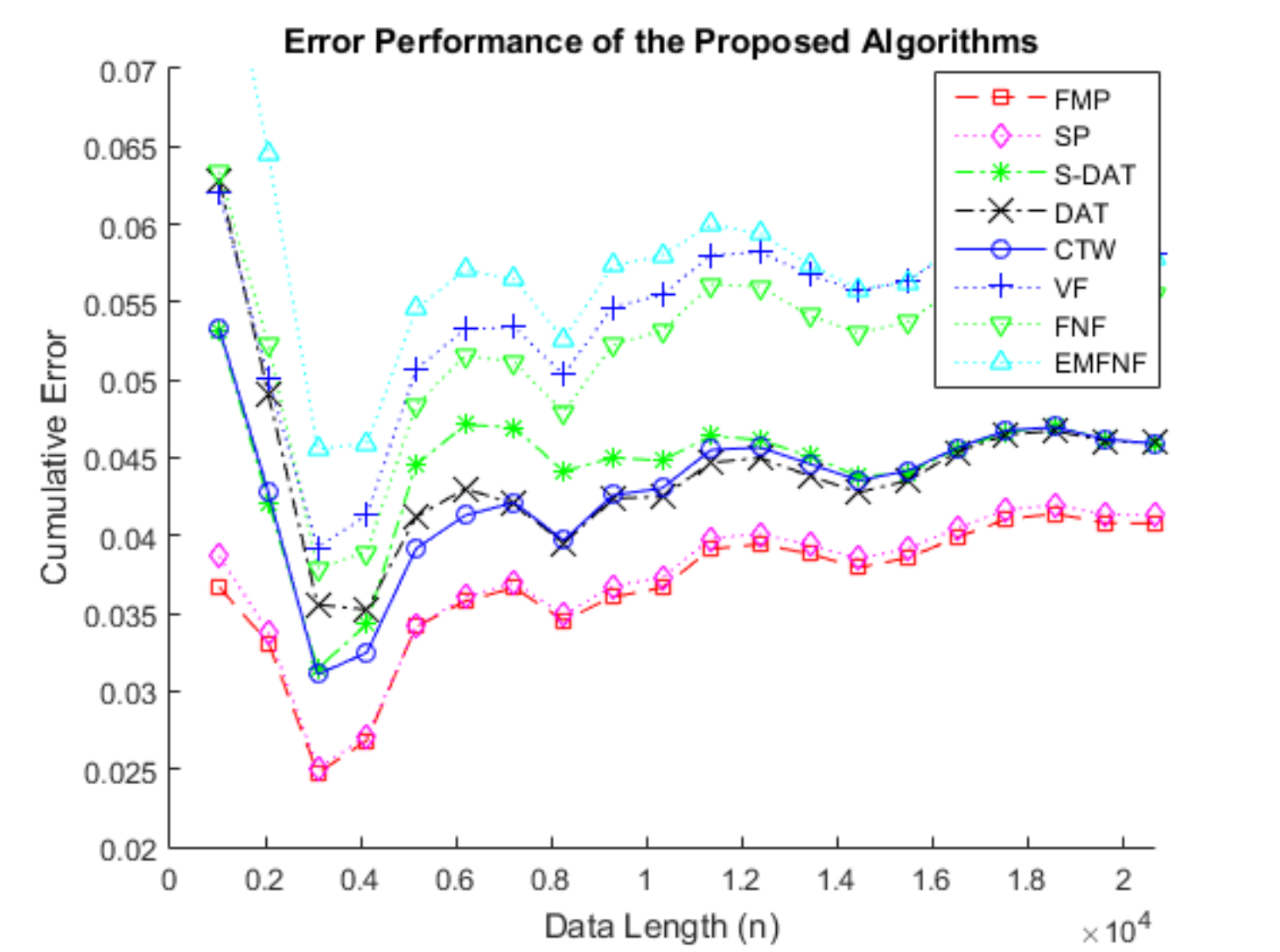} 
	\caption{Time accumulated error performances of the proposed algorithms for California Housing Data Set. \label{fig:california}}
\end{figure}

In this subsection, we mainly focus on assessing the merits of our algorithms. We first consider the regression of a benchmark real-life problem that can be found in many data set repositories such as: California Housing, which is an $m = 8$ dimensional database consisting of the estimations of median house prices in the California area \cite{ltorgo}. There exist more than 20000 data samples for this dataset. For this experiment, we set the learning rates to 0.004 for FMP and SP, 0.01 for the S-DAT and the DAT, 0.02 for the CTW, 0.05 for the VF, 0.005 for the FNF and the EMFNF. Fig. \ref{fig:california} illustrates the normalized time accumulated error rates of the stated algorithms. We emphasize that the FMP and the SP significantly outperforms the state of the art.\par
\begin{figure}[t]
	\centering
	\includegraphics[scale=0.35]{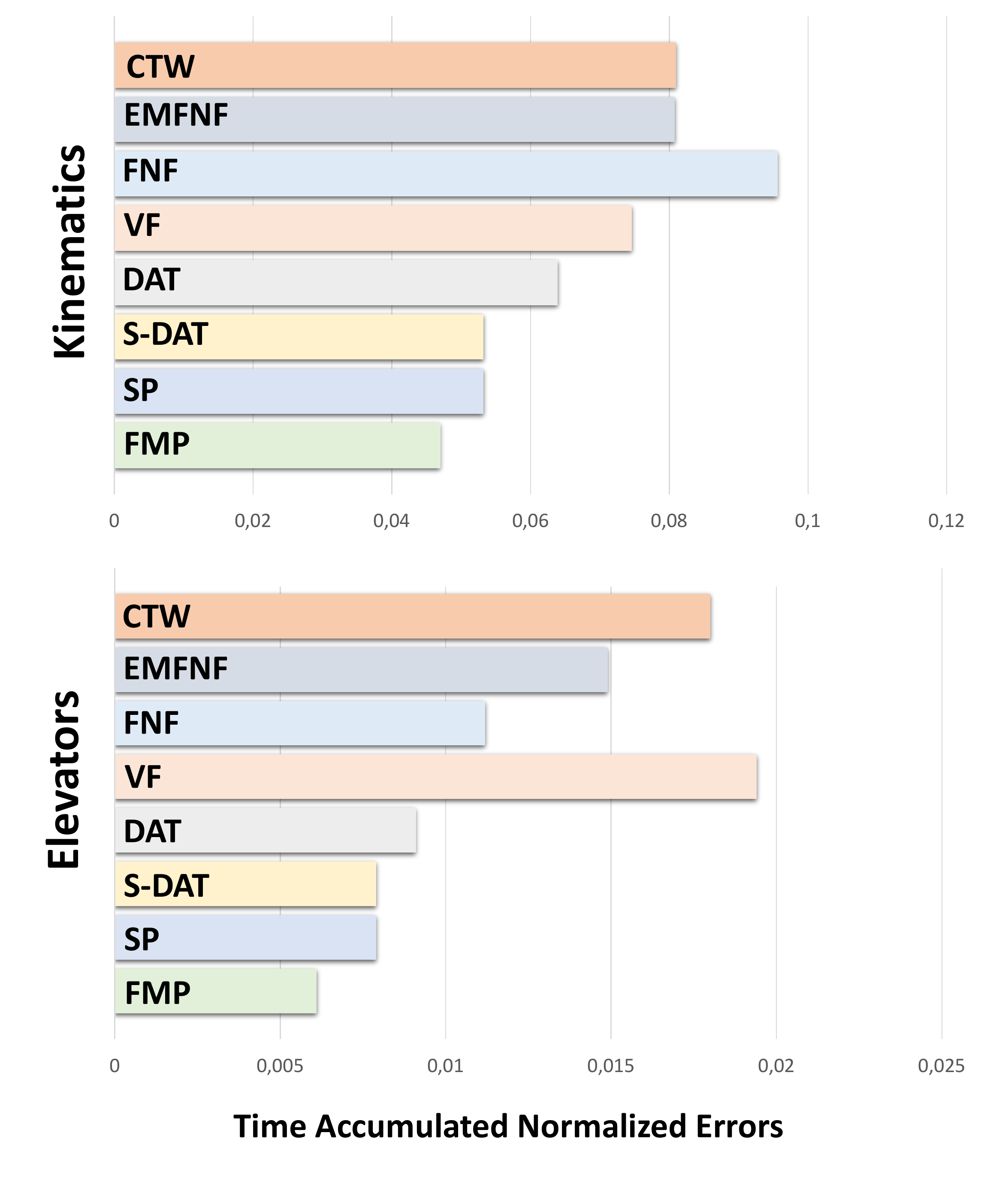} 
	\caption{Time accumulated error performances of the proposed algorithms for Kinematics and Elevators Data Set. \label{fig:realDatErr}}
\end{figure}

We also consider two more real and synthetic data sets. The first one is Kinematics, which is an $m = 8$ dimensional dataset where a realistic simulation of an 8 link robot arm is performed \cite{ltorgo}. The task is to predict the distance of the end-effector from a target. There exist more than 50000 data samples. The second one is Elevators, which has an $m = 16$ dimensional data sequence obtained from the task of controlling an F16 aircraft\cite{ltorgo}. This dataset provides more than 50000 samples. In Fig. \ref{fig:realDatErr}, we present the steady state error performances of the proposed algorithms. We emphasize that our algorithms achieve considerably better performance compared to the others for both datasets. \par
\begin{figure}[t]
	\centering
	\includegraphics[scale=0.4]{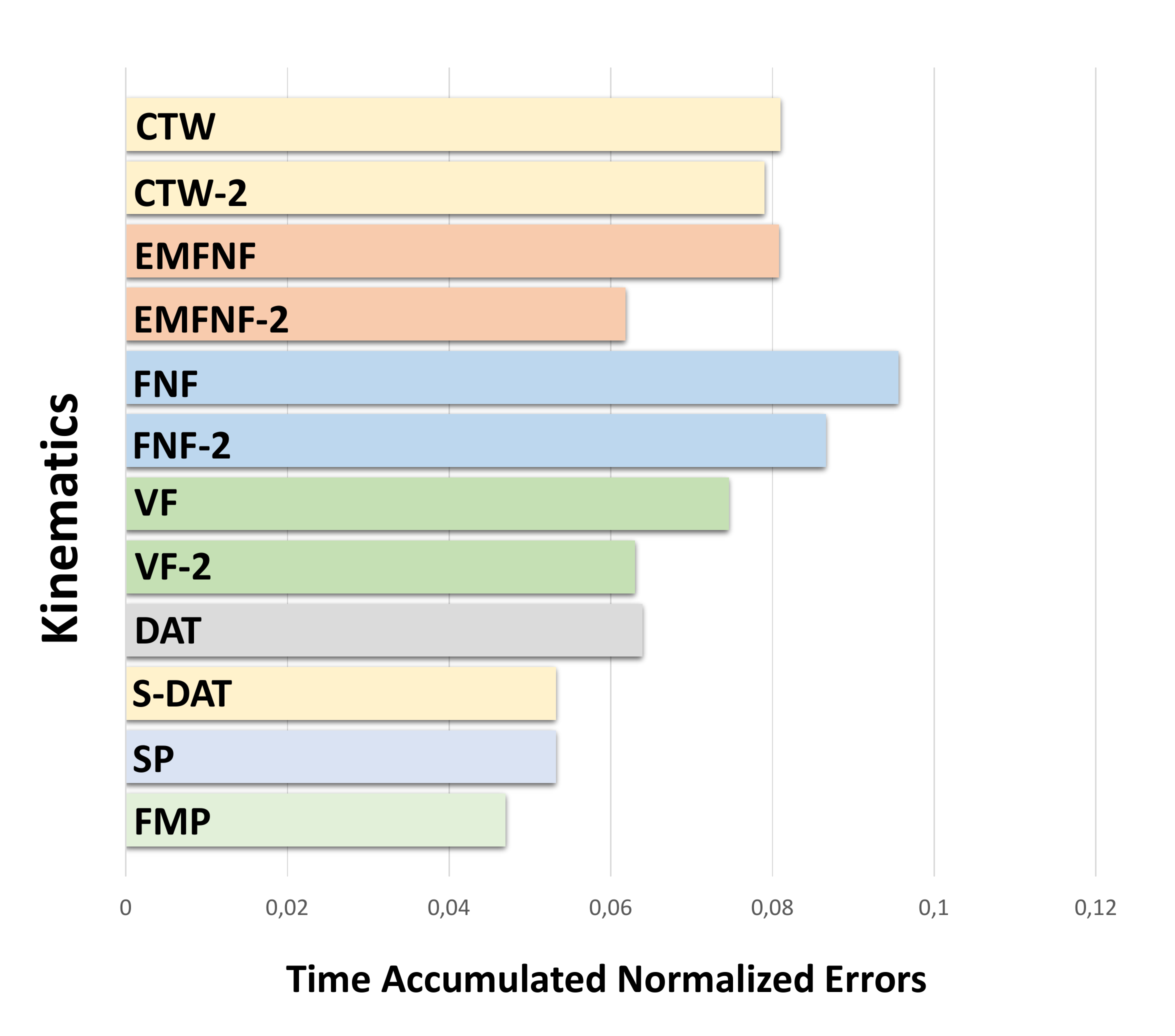} 
	\caption{Time accumulated error performances of the proposed algorithms for Kinematics Data Set. The second order adaptation methods are used for all algorithms. \label{fig:RealDataErrors2}}
\end{figure}

Special to this subsection, we perform an additional experiment using the Kinematics dataset to illustrate the effect of using second order methods for the adaptation. Usually, algorithms like CTW, FNF, EMFNF, VF and DAT use the gradient based first order methods for the adaptation algorithm due to their low computational demand. Here, we modified the adaptation part of these algorithms and use the second order Newton-Raphson methods instead. In Fig. \ref{fig:RealDataErrors2}, we illustrate a comparison that involves the final error rates of both the modified and the original algorithms. We also keep our algorithms in their original settings to demonstrate the effect of using piecewise linear functions when the same adaptation algorithm is used. In Fig. \ref{fig:RealDataErrors2}, the CTW-2, the EMFNF-2, the FNF-2 and the VF-2 state for the algorithms using the second order methods for the adaptation. The presented S-DAT algorithm already corresponds to the DAT algorithm with the second order adaptation methods. Even though this modification decreases the final error of all algorithms, our algorithms still outperform their competitors. Additionally, in terms of the computational complexity, the algorithms EMFNF-2, FNF-2 and VF-2 become more costly compared to the proposed algorithms since they now use the second order methods for the adaptation. There exist only one algorithm, i.e., CTW-2, that is more efficient, but it does not achieve a significant gain on the error performance.

\subsection{Chaotic Signals}\label{sec:chaoticSig}     
Finally, we examine the error performance of our algorithms when the desired data sequence is generated using chaotic processes, e.g. the Gauss map and the Lorenz attractor. We first consider the case where the data is generated using the Gauss map, i.e.,
\begin{equation}
y_{t} = \exp\big(-\alpha x_{t}^{2}\big) + \beta
\end{equation}
which exhibits a chaotic behavior for $\alpha = 4$ and $\beta = 0.5$. The desired data sequence is represented by $y_t$ and $x_t \in \mathbbm{R}$ corresponds to $y_{t-1}$. $x_{0}$ is a sample from a Gaussian process with zero-mean and unit variance. The learning rates are set to 0.004 for the FMP, 0.04 for the SP, 0.05 for the S-DAT and the DAT, 0.025 for the VF, the FNF, the EMFNF and the CTW.\par
\begin{figure}
	\centering
	\includegraphics[scale=0.7]{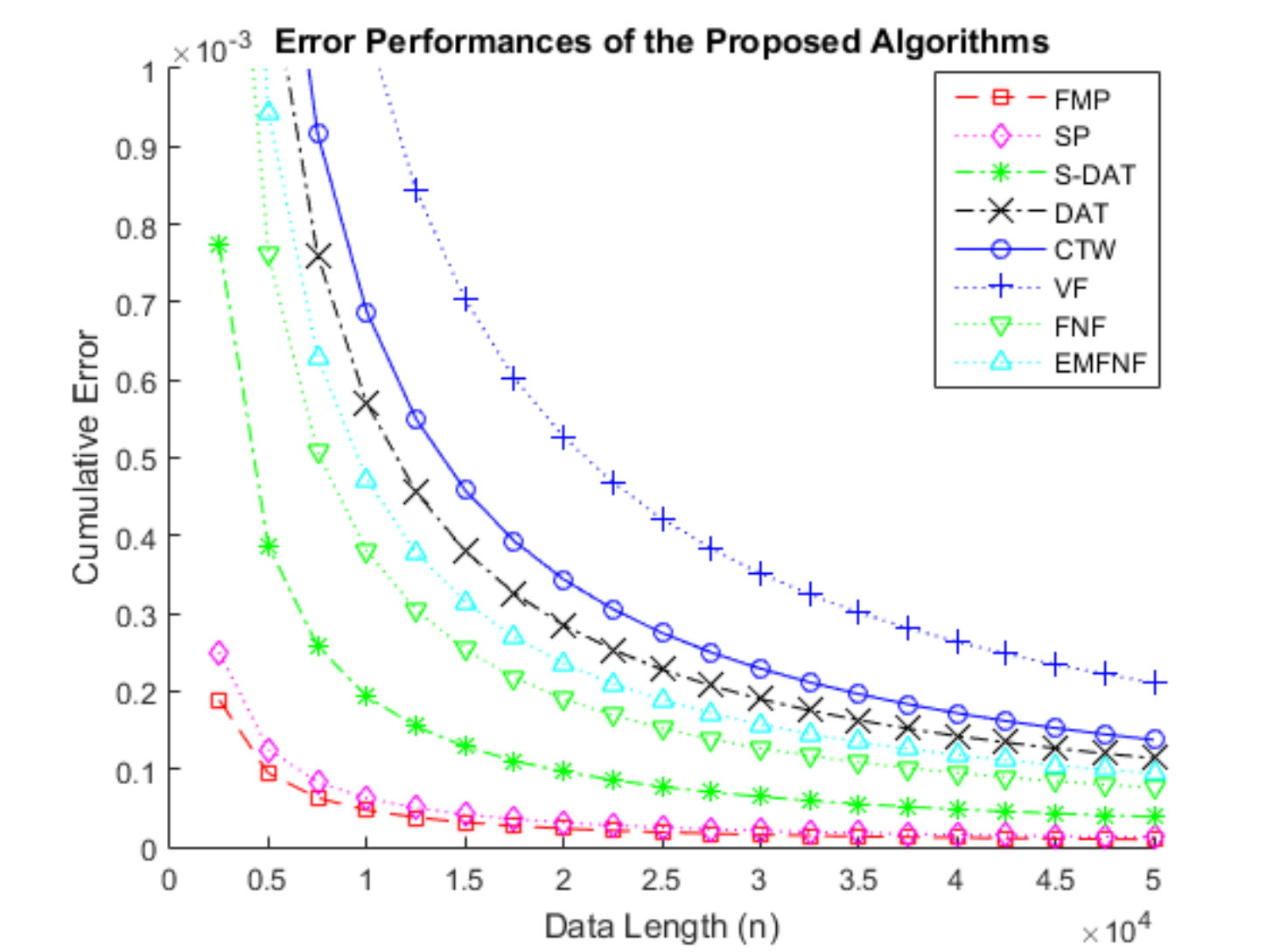} 
	\caption{Regression Error Rates for the Gauss map.\label{fig:chaotic_gauss}}
\end{figure}
As the second experiment, we consider a scenario where we use a chaotic signal that is generated from the Lorenz attractor, which is a set of chaotic solutions for the Lorenz system. Hence, the desired signal $y_t$ is modeled by
\begin{align}
y_t &= y_{t-1} + (\sigma(u_{t-1}-y_{t-1}))dt \\
u_t &= u_{t-1} + (y_{t-1}(\rho - v_{t-1})-u_{t-1})dt \\
v_t &= v_{t-1} + (y_{t-1}u_{t-1}- \beta v_{t-1})dt,
\end{align}
where $\beta = 8/3$, $\sigma = 10$, $\rho = 28$ and $dt = 0.01$. Here, $u_t$ and $v_t$ are used to represent the two dimensional regression space, i.e., the data vector is formed as $\vec{x}_t = [u_t,v_t]^T$. We set the learning rates to 0.005 for the FMP, 0.006 for the SP, 0.0125 for the S-DAT, 0.01 for the DAT, the VF, the FNF, the EMFNF and the CTW.\par
\begin{figure}
	\centering
	\includegraphics[scale=0.7]{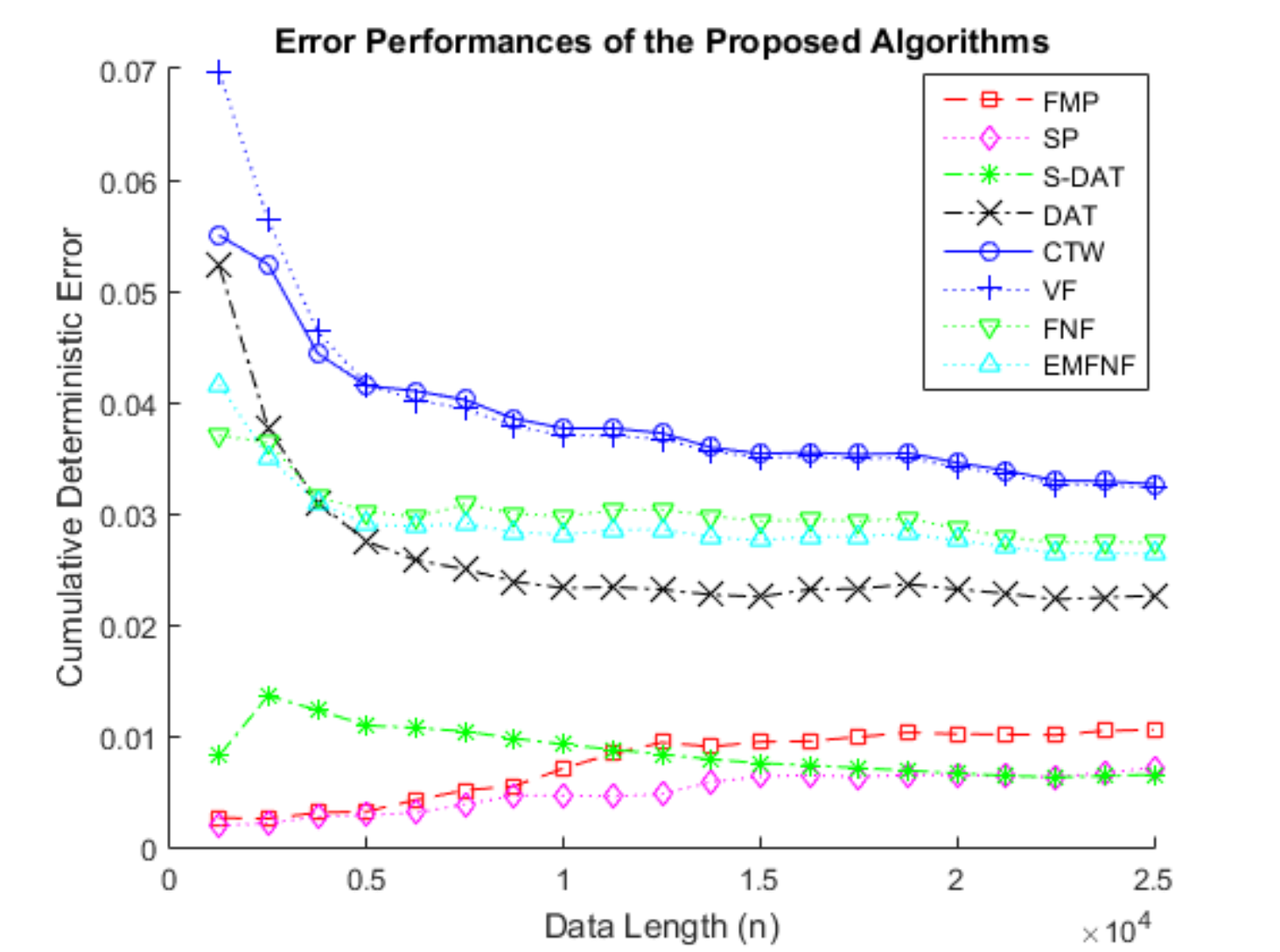} 
	\caption{Regression Error Rates for the Lorenz attractor.\label{fig:chaotic_lorenz}}
\end{figure}

In Fig. \ref{fig:chaotic_gauss} and \ref{fig:chaotic_lorenz}, we represent the error performance of the proposed algorithms for the Gauss map and the Lorenz attractor cases respectively. In both cases, the proposed algorithms attain substantially faster convergence rate and better steady state error performance compared to the state of the art. Even for the Lorenz attractor case, where the desired signal has a dependence on more than one past output samples, our algorithms outperform the competitors.
\section{Concluding Remarks}\label{sec:conclusion}
In this paper, we introduce three different highly efficient and effective nonlinear regression algorithms for online learning problems suitable for real life applications. We process only the currently available data for regression and then discard it, i.e., there is no need for storage. For nonlinear modeling, we use piecewise linear models, where we partition the regressor space using linear separators and fit linear regressors to each partition. We construct our algorithms based on two different approaches for the partitioning of the space of the regressors. As the first time in the literature, we adaptively update both the region boundaries and the linear regressors in each region using the second order methods, i.e., Newton-Raphson Methods. We illustrate that the proposed algorithms attain outstanding performance compared to the state of art even for the highly nonlinear data models. We also provide the individual sequence results demonstrating the guaranteed regret performance of the introduced algorithms without any statistical assumptions.   

\section*{Acknowledgment}
This work is supported in part by Turkish Academy of Sciences Outstanding Researcher Programme, TUBITAK Contract No. 113E517, and Turk Telekom Communications Services Incorporated.

\bibliographystyle{elsarticle-num}
\bibliography{msaf_references}

\end{document}